\documentclass[lettersize,journal]{IEEEtran}
\usepackage{amsmath,amsfonts}
\usepackage{algorithm, algorithmicx, algpseudocode}
\usepackage{array}

\usepackage{mathrsfs}
\usepackage{stfloats}

\usepackage{booktabs}
\usepackage{multirow}
\usepackage{longtable}
\usepackage {rotating} 
\usepackage{subfigure}
\usepackage[caption=false,font=normalsize,labelfont=sf,textfont=sf]{subfig}
\usepackage{textcomp}
\usepackage{stfloats}
\usepackage{url}
\usepackage{verbatim}
\usepackage{graphicx}
\usepackage{balance}
\usepackage{cite}
\usepackage{cuted}
\def\BibTeX{{\rm B\kern-.05em{\sc i\kern-.025em b}\kern-.08em
		T\kern-.1667em\lower.7ex\hbox{E}\kern-.125emX}}
	
\begin{document}
	
	\title{An Adaptive Balance Search  Based Complementary Heterogeneous Particle Swarm Optimization Architecture }
	
	\author{
\textbf{	This work has been submitted to the IEEE for possible publication. Copyright may be transferred without notice, after which this version may no longer be accessible.}\\
Zhenxing Zhang, Tianxian Zhang, Xiangliang Xu, Lingjiang Kong

}

	\markboth{Journal of \LaTeX\ Class Files,~Vol.~18, No.~9, April~2024}%
	{Shell \MakeLowercase{\textit{et al.}}: A Sample Article Using IEEEtran.cls for IEEE Journals}
	
	
	\maketitle
	
	\begin{abstract}
		
A series of modified cognitive-only particle swarm optimization (PSO) algorithms effectively mitigate premature convergence by constructing distinct vectors for different particles. However, the underutilization of these constructed vectors hampers convergence accuracy. In this paper, an adaptive balance search based complementary heterogeneous PSO architecture is proposed, which consists of a complementary heterogeneous PSO (CH\textit{x}PSO) framework and an adaptive balance search (ABS) strategy. The CH\textit{x}PSO framework mainly includes two update channels and two subswarms. Two channels exhibit nearly heterogeneous properties while sharing a common constructed vector. This ensures that one constructed vector is utilized across both heterogeneous update mechanisms. The two subswarms work within their respective channels during the evolutionary process, preventing interference between the two channels. The ABS strategy precisely controls the proportion of particles involved in the evolution in the two channels, and thereby guarantees the flexible utilization of the constructed vectors, based on the evolutionary process and the interactions with the problem’s fitness landscape. Together, our architecture ensures the effective utilization of the constructed vectors by emphasizing exploration in the early evolutionary process while exploitation in the later, enhancing the performance of a series of modified cognitive-only PSOs. Extensive experimental results demonstrate the generalization performance of our architecture.
		
	\end{abstract}
	
	\begin{IEEEkeywords}
		 Heterogeneity,  problem’s fitness landscape, exploration and exploitation, proportion of particles, particle swarm optimization (PSO).

	\end{IEEEkeywords}
	
	\section{Introduction} \label{Introduction}
	\IEEEPARstart{I}{n real} world, many engineering applications can be modeled as optimization problems with characteristics like discontinuity, multimodality, and non-differentiability \cite{ wang2022stochastic, isho2020persistence, alrashidi2008survey, mistry2016micro}. To address these challenges, various population-based optimization algorithms have been developed, such as particle swarm optimization (PSO) \cite{kennedy1995particle}, ant colony optimization \cite{7511696}, and genetic algorithm \cite{katoch2021review}. Among these, PSO has garnered significant attention due to its simple optimization mechanism and effective optimization performance \cite{suganthan1999particle,shi1998modified, zhang2007hybrid, vlachogiannis2006comparative}.
	
	In fact, when solving multimodal problems, particles in standard PSO, especially in global version PSO (\textit{global} PSO), are often captured prematurely by local attractors, leading to premature convergence \cite{li2021adaptive, zhang2021promotive, bangyal2021new}. To address this, an extreme PSO variant, known as cognitive-only PSO \cite {kennedy1997particle}, was proposed. Mechanistically, cognitive-only PSO ignores the influence of neighbors, with each particle guided solely by its personal best vector. This allows each particle to search independently, making particles less likely to converge prematurely toward local attractors. Although this update mechanism promotes exploration and helps mitigate premature convergence, it comes at the cost of reduced convergence accuracy. Kennedy \cite{kennedy1997particle} argued that the problem was one of failure to find an optimal region. A major reason is that the personal best vectors contain only personal knowledge, guiding particles to tend to search the regions that they have been initialized, and preventing them from moving into the optimal regions.

	In response, researchers have enriched the information within the personal best vectors by introducing various topologies \cite{mendes2004fully, blackwell2018impact}, learning strategies \cite{liang2006comprehensive, 7271066, zhan2009orthogonal}, and other methods \cite{li2015composite, ren2013scatter}. These efforts have led to various modified cognitive-only PSOs\footnote{Cognitive-only PSO can be viewed as a special form.} that both mitigate premature convergence and improve convergence accuracy. For instance, Mendes \textit{et al.} \cite{ mendes2004fully} constructed a new vector $ \vec {P}m$ for \textit{m}th particle by incorporating all personal best information of its neighbors. This method helps the particles avoid being prematurely captured by local attractors and find optimal regions. However, particles may oscillate between peaks if neighbors are from different ecological niches, which hampers both exploration and exploitation. To address this, Qu \textit{et al.} \cite{qu2012distance} further selected the neighbors from the same ecological niche based on Euclidean distance, constructing new vectors $\boldsymbol{P}_i$.
	Moreover, the size of neighbors is dynamically adjusted to finely control the information, pursuing the trade-off between exploration and exploitation. However, it neglects the case that a particle that has discovered the region corresponding to the global optimum in some dimensions may have a low fitness value because of the poor solutions in the other dimensions. For this reason, Liang \textit{et al.} \cite{liang2006comprehensive} proposed a comprehensive learning (CL) strategy, which enriches personal best vectors by utilizing information from the entire swarm across all dimensions, constructing new vectors $pbest_{fi}$. Besides, learning probability is introduced to allow particles to have different levels of exploration and exploitation. 
	
	Overall, these modified cognitive-only PSOs have made significant contributions to both avoiding premature convergence and enhancing convergence accuracy. On one hand, the update mechanism in these algorithms allows each particle to be guided by one distinct vector, showing the property that the mechanism prefers guiding the swarm exploration, and thereby effectively alleviating premature convergence. On the other hand, and more crucially, these algorithms incorporate information from other particles when constructing the vector for one particle. This ensures that the constructed vectors have the strong ability to guide the particles to explore and exploit, helping find optimal regions, and thereby helping convergence accuracy. Nevertheless, there is still room for improvement, which is mainly due to two common issues:

\textbf{\textit{Common issue} 1:} \textbf{The abilities of constructed vectors to guide particle exploration and exploitation cannot be fully utilized through one single update mechanism.} In these modified cognitive-only PSOs, the constructed vectors typically operate within the confines of one single update mechanism, which prefers guiding the swarm exploration. The ability of constructed vectors to guide particle exploration is effectively utilized by aligning with the mechanism’s property, while the ability to guide particle exploitation may be partially neglected due to a weaker alignment with its property. This issue becomes particularly pronounced in the later evolutionary process, where it can prevent the particles from clustering toward the global optimal region.

\textbf{\textit{Common issue} 2:} \textbf{The abilities of constructed vectors to guide particle exploration and exploitation cannot be flexibly utilized through one single fixed upper bound.} In these modified cognitive-only PSOs, the fixed vectors are commonly reutilized within a predetermined number of iterations, here referred to as the upper bound. For example, in FIPS \cite{ mendes2004fully}, the upper bound is set to one because FIPS reconstructs vectors before each iteration starts. In CLPSO \cite{liang2006comprehensive}, the upper bound is experimentally set to seven. The single fixed upper bound inevitably causes the same upper bound when the two abilities of constructed vectors are utilized, thereby resulting in an equal emphasis on the utilization of the two abilities throughout the evolutionary process. This leads to an overemphasis on exploration while underemphasis on exploitation, particularly in the later evolutionary process. It hinders the trend of exploration in the early evolutionary process while exploitation in the later \cite{vcrepinvsek2013exploration}. 
  
    \subsection{Research Motivations}
The development of heterogeneity \cite{du2016heterogeneous}, particularly update-rule heterogeneity \cite{de2009heterogeneous}, provides ideas to address \textbf{\textit{Common issue} 1}. Update-rule heterogeneity, as defined by De Oca \textit{et al.} \cite{de2009heterogeneous}, allows for the integration of two or more update mechanisms within a single algorithm, enabling particles to search in different (but complementary) ways. This has driven various heterogeneous PSOs \cite{chu2014ahps2, du2019network, zheng2016hybrid, lin2023heterogeneous, wang2020heterogeneous}. A notable example is HCLPSO \cite{lynn2015heterogeneous}, which enhances the traditional CLPSO (CLPSO prefers guiding the swarm exploration) by incorporating an update mechanism that prefers guiding the swarm exploitation. This complementary update mechanism is a modified version of \textit{global} PSO, where the CL strategy constructs new vectors to replace personal best vectors but retains global best positions. HCLPSO also features two distinct subswarms: one for exploration, which follows the traditional CLPSO update mechanism, and another for exploitation, which operates under the update mechanism of modified \textit{global} PSO. In the former, vectors are constructed using information from the exploration subswarm, while in the latter, the vectors' information is from both subswarms. It is clear that it avoids the exploration subswarm being influenced by the exploitation subswarm, and the vectors constructed in the two update mechanisms are different. Despite these advancements, the algorithm still suffers from the same issues: a single fixed constructed vector is used only in one update mechanism and is restricted by a single fixed upper bound.

To address the \textbf{\textit{common issue} 1}, if we embed the vectors constructed by one modified cognitive-only PSO simultaneously into another update mechanism that prefers guiding the swarm exploitation, we can ensure that the ability of the constructed vectors to guide particle exploitation is aligned with the property of the complementary update mechanism, preventing it from being partially neglected. 

To address the \textbf{\textit{common issue} 2}, if we set two different upper bounds for each ability based on the traditional upper bound, and let the two upper bounds be dynamically adjusted with the evolutionary process, we may avoid an equal emphasis on two abilities of constructed vectors. Besides, if we can precisely determine which ability to utilize in each iteration by the interactions with the evolutionary environment, then we may be able to flexibly utilize the constructed vectors.

\subsection{Research Contributions}

We propose an Adaptive Balance Search Based Complementary Heterogeneous Particle Swarm Optimization (CH\textit{x}PSO-ABS\footnote{\textit{x} identifies the generalization of the architecture, specifically represents the vector-construction methods in the modified cognitive-only PSOs.}) architecture. The architecture can embed a series of vector-construction methods from modified cognitive-only PSOs to improve their performance. Our contributions are as follows: 

\begin{enumerate}{}{}
	\item{ We introduce a Complementary Heterogeneous Particle Swarm Optimization (CH\textit{x}PSO) framework mainly with two update channels and two subswarms. Two channels are nearly heterogeneous for guiding particles to explore or exploit, respectively, while both of them contain the same vector, which is constructed by modified cognitive-only PSOs. This guarantees that each of the abilities of the constructed vectors is matched with the properties of the update channels. Two subswarms work in their respective channels, ensuring that the two abilities of constructed vectors do not interfere with each other when utilized.}
	
	\item{ We propose an Adaptive Balance Search (ABS) strategy including a cap limiter, an R \& P box, and an adaptive selector. The cap limiter adaptively governs the two upper thresholds of utilizing two abilities, according to the number of function evaluations. The R $\&$ P box regulates the number of utilization of the two abilities, according to the updated states of particles. The adaptive selector controls the proportion of particles involved in the evolution in the two channels, thereby effectively utilizing the constructed vectors.}
	
	\item{ We embed the vector-construction methods of cognitive-only PSO and CLPSO into our architecture and conduct extensive experiments to verify the effectiveness of our architecture.}
	
\end{enumerate}

The remainder of this paper is organized as follows: Section \ref{Related work} reviews the related work. The generic architecture CH\textit{x}PSO-ABS is proposed in Section \ref{DCPSO-ABS}. Section \ref{experiment} encompasses the experiments along with discussions. The conclusion is drawn in Section \ref{conclusion}.

\begin{figure*}[!t]
	\centering
	\includegraphics[width=6.3in]{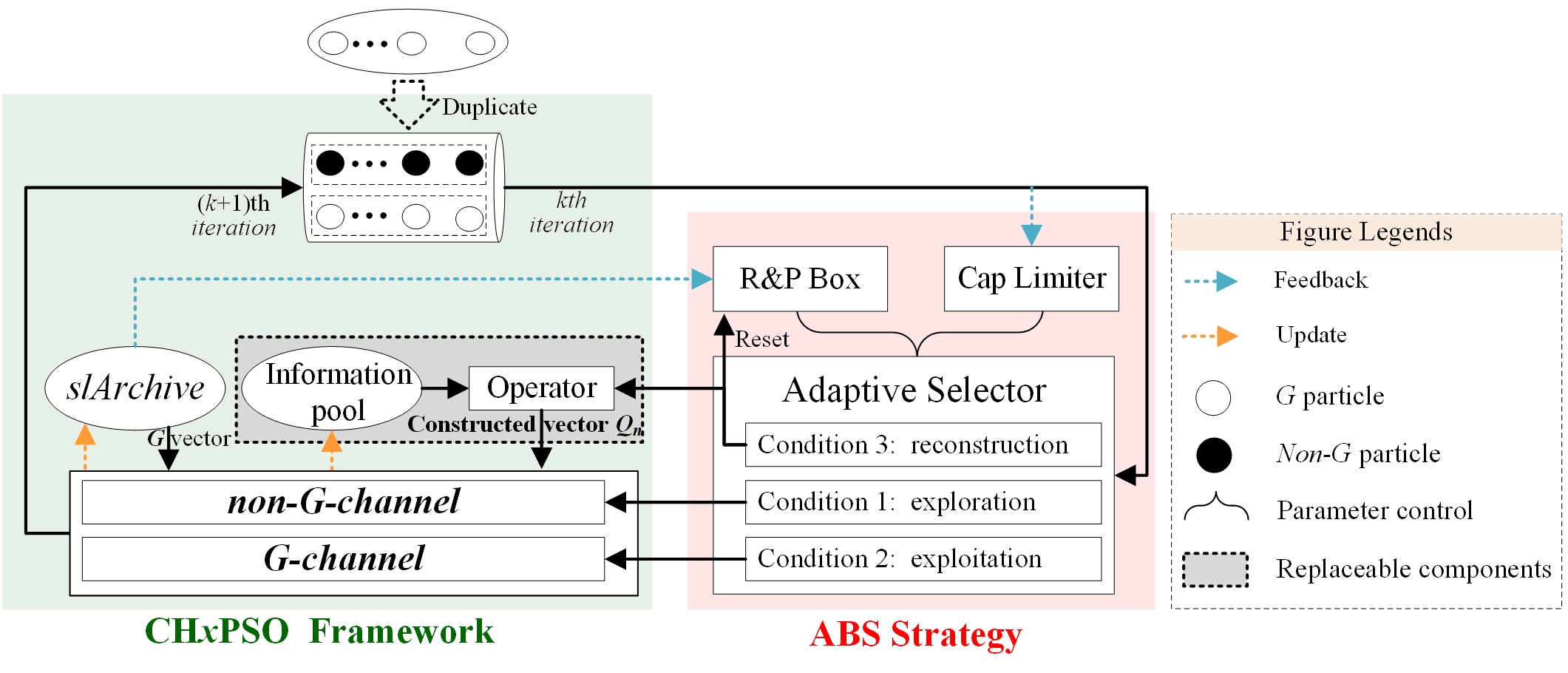}
	\caption{Generic architecture CH\textit{x}PSO-ABS.}
	\label{generic architecture}
\end{figure*}

\section{Related Work}\label{Related work}

\subsection{\textit{global} PSO}
 In \textit{global} PSO, each particle is to move towards two positions, the best position found by the particle and the best position found by the entire swarm.  For a $D$-dimensional optimization problem, the flight of the $n$th particle is controlled by previous velocity  $\boldsymbol{V_n }= (v^1_n,v^2_n,...,v^D_n)$, previous position $ \boldsymbol{X_n} = (x^1_n,x^2_n,...,x^D_n)$, personal best vector $\boldsymbol{P_n} = (p^1_n,p^2_n,...,p^D_n)$ and global best vector $ \boldsymbol{G}=(g^1,g^2,...,g^D)$, which can be specifically described by 
 
 \begin{equation} \label{gbest PSO}
 	\begin{cases}
 		{v}^d_n = w\cdot {v}^d_n	+{c_1}\cdot {r^d_1} \cdot ({p}^d_n-{x}^d_n)	+{c_2}\cdot {r^d_2} \cdot (g^d-{x}^d_n)\\ 
 		{x}^d_n={x}^d_n+{v}^d_n
 	\end{cases},
 \end{equation}
 where $w$ represents the inertia weight,   and $ {c_1}$ and $ {c_2}$ represent the cognitive and social acceleration coefficients, respectively. $r^d_1$ and $r^d_2$ are two random numbers selected evenly within [0,1]. Lynn \textit{et al.} \cite{lynn2015heterogeneous} claimed that the exploitation can be enhanced by appropriately setting $w$ to a linearly decreasing function of the number of iterations in the range of $0.99-0.2$, and employing time-varying acceleration coefficients $c_1=2.5-0.5$ and $c_2=0.5-2.5$\cite{ratnaweera2004self}.

\subsection{Cognitive-Only PSO}
The cognitive-only PSO \cite{kennedy1997particle}, obtained by removing the social component from Eq. (\ref{gbest PSO}), is given as follows:
\begin{equation} \label{Cognition-only PSO}
	\begin{cases}
		{v}^d_n = w\cdot {v}^d_n	+{c_1}\cdot {r^d_1} \cdot ({p}^d_n-{x}^d_n)\\ 
		{x}^d_n={x}^d_n+{v}^d_n
	\end{cases}.
\end{equation}
This model	results in more exploration because each particle becomes a climber \cite{engelbrecht2010heterogeneous}. Nevertheless, the particle would search in a small region around its own position when $\boldsymbol{P_n} = \boldsymbol{X_n}$.

\subsection{CLPSO}
	
	CLPSO \cite{liang2006comprehensive} is proposed to address the premature convergence due to dramatic loss of diversity. Based on cognitive-only PSO, this algorithm adopts a CL strategy to construct  vectors, $\textbf{\textit{P}}_{\textbf{\textit{f}}n} = (p^1_{fn(1)},...,p^D_{fn(D)})$, which are informed by the  personal best positions of different particles with a learning probability $Pc_n$. The $Pc_n$ is calculated by
	
			\begin{equation} \label{CLPSO}
		Pc_n = 0.05 + 0.45 \cdot \dfrac{\exp(\dfrac{10(n-1)}{N-1})-1}{\exp(10)-1}
		\\ ,
	\end{equation}
where \textit{N} denotes the total number of particles. The  velocity update equation  is 
	\begin{equation} \label{CLPSO}
		{v}^d_n = w\cdot {v}^d_n+{c}\cdot {r^d} \cdot ({p^d_{fn(d)}}-{x}^d_n)\\ ,
	\end{equation}
	where $fn(d)$ defines which particle's personal best vector  that $n$th particle should follow on $d$-dimension. Moreover, to ensure that particles learn from good exemplars and to minimize the time wasted in bad directions, the authors allow particles to learn from exemplars until the particle stops improving for a certain number of generations, called the refreshing gap. The refreshing gap is experimentally set to seven for all tested functions in \cite{liang2006comprehensive}.

\section{Generic Architecture CH\textit{x}PSO-ABS}\label{DCPSO-ABS}

 This section introduces a generic architecture CH\textit{x}PSO-ABS to enhance the performance of a series of modified cognitive-only PSOs.

    \subsection{Overall Description}\label{Overall Description}
The generic architecture of CH\textit{x}PSO-ABS is illustrated in Fig. \ref{generic architecture}. It consists of two main components: CH\textit{x}PSO framework and ABS strategy. 

 The CH\textit{x}PSO framework includes two update channels, two subswarms, a \textit{slArchive}, an information pool, and an operator. Specifically, the two update channels with nearly heterogeneous properties share a common constructed vector, $\boldsymbol{Q_n}$, ensuring that the two abilities of the constructed vector align with their properties. The two subswarms, each assigned to a respective channel throughout the evolutionary process, ensure that the two abilities of the constructed vectors do not interfere with each other when utilized. The \textit{slArchive} stores the single-layer best vectors (a new term introduced in this paper, detailed in Section \ref{slArchive}), providing a uniform standard for utilizing the constructed vectors. The information pool and operator are conceptual components (replaceable components) that can be specified according to different modified cognitive-only PSOs, for storing key information and constructing vectors.
			\begin{figure}[!t]
	\centering
	\includegraphics[width=3in]{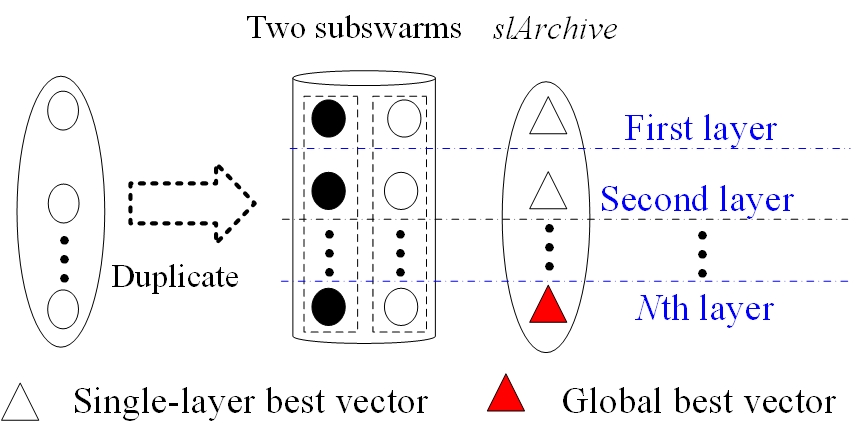}
	\caption{The correspondence between initial swarm, subswarms and \textit{slArchive}.}
	\label{Correspondence}
\end{figure}

The ABS strategy includes a cap limiter, a reward and punishment box (R $\&$ P box), and an adaptive selector. Specifically, the cap limiter adaptively governs the two upper thresholds for the utilization of two abilities of each constructed vector according to the number of function evaluations. The R \& P box regulates the number of utilization of the constructed vectors, according to the updated states of single-layer best vectors. The adaptive selector, stimulated by the cap limiter and R $\&$ P box, decides which ability of the constructed vector should be utilized, or to reconstruct the vector.  

   All components work in an orderly manner, improving the performance of a series of modified cognitive-only PSOs. We provide a detailed description of how the architecture works. Specifically, in the \textit{k}th iteration, firstly, all particles in the two subswarms are prepared as candidates for the current evolutionary stage. Then, the adaptive selector selects the appropriate condition to execute. If condition 1 is executed, the adaptive selector sends the \textit{non-G} particles into the \textit{non-G-channel} to utilize the ability of constructed vectors to guide particle exploration. The particles' positions, information pool, and \textit{slArchive} are updated. The updated results in the \textit{slArchive} are fed back to the R $\&$ P box.  If condition 2 is executed, the adaptive selector will select the \textit{G} particles into the \textit{G-channel} to utilize the ability of constructed vectors to guide particle exploitation, following the same process as condition 1. Besides, if neither condition 1 nor condition 2 is satisfied, then condition 3 is executed. The  operator extracts the relevant information from the information pool for reconstructing  $\boldsymbol{Q_n}$ vector, and the adaptive selector resets the relevant parameters in the R $\&$ P box to  self-drive the execution of condition 1.   Finally, if the stop condition is not met, the architecture proceeds to the $(k+1)$th iteration, and the number of function evaluations is fed back to the cap limiter.

	\newcounter{TempEqCnt}
	\setcounter{TempEqCnt}{5}
	\setcounter{equation}{4}
	
	\begin{figure*}[b]
		\hrulefill
		\begin{align}
			\boldsymbol{L_n}(k+1)=
			\begin{gathered} 
				\left\{ \begin{gathered}
					\underset{non-G}{\boldsymbol{X_n}}(k+1)*(f(	\underset{non-G}{\boldsymbol{X_n}}(k+1)) \leq f(\boldsymbol{L_n}(k)))+\boldsymbol{L_n}(k)*(f(\underset{non-G}{\boldsymbol{X_n}}(k+1))>f(\boldsymbol{L_n}(k)))\\
					$\textit{or}$\\				
					\underset{G}{\boldsymbol{X_n}}(k+1)*(f(\underset{G}{\boldsymbol{X_n}}(k+1))\leq f(\boldsymbol{L_n}(k)))+\boldsymbol{L_n}(k)*(f(\underset{G}{\boldsymbol{X_n}}(k+1))>f(\boldsymbol{L_n}(k)))\\
				\end{gathered}  \right. \hfill \\
			\end{gathered}
			\label{pbest update formular}
		\end{align}
	\end{figure*}
	\setcounter{equation}{\value{TempEqCnt}}

		\subsection{CHxPSO Framework}\label{CHxPSO Framework}	

The framework consists of five components: two update channels, two subswarms, the \textit{slArchive}, the information pool, and the operator.  These components are described in detail below:
	
	\textit{\textbf{1) Two subswarms:}}\label{Two subswarms}  We employ two subswarms, named \textit{non-G} subswarm and \textit{G} subswarm, with their particles referred to as \textit{non-G}  and \textit{G} particles, respectively. Unlike the common approach of generating subswarms through random initialization \cite{li2021adaptive,  lynn2015heterogeneous}, we create two subswarms by duplication. Specifically, we generate a swarm using traditional initialization and then create a second swarm by duplicating the first one. As expected, both subswarms contain the same information. Therefore, the assignment of subswarms to two update channels can be made arbitrarily before the evolutionary process starts. Moreover, we set that the \textit{non-G} subswarm works in \textit{non-G-channel}  and \textit{G} subswarm works in \textit{G-channel}, through the whole evolutionary process. Therefore, it ensures that the two abilities of the constructed vector do not interfere with each other when utilized.

\begin{figure} [!]
	\subfigure[]{
		\includegraphics[width=1.67in]{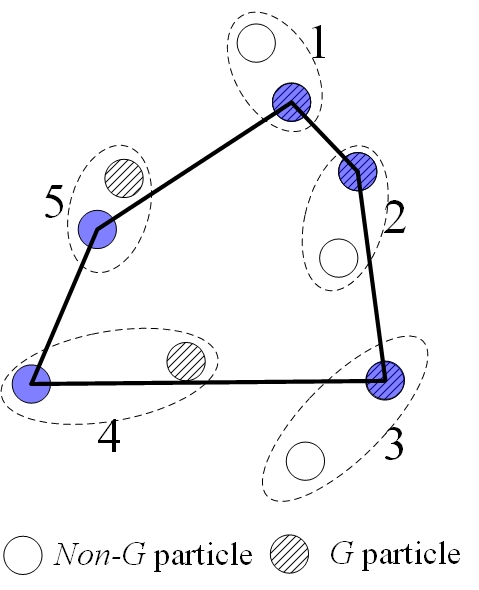}\\
	}%
	\hspace{-2mm}
	\subfigure[]{
		\includegraphics[width=1.7in]{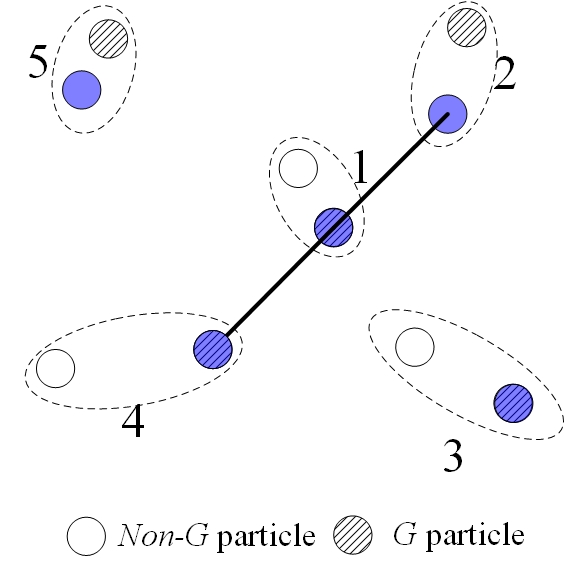}\\
	}
	
	\caption{
		Schematic diagram of topology construction and exemplar selection in our  architecture. The blue-labeled circles represent active particles. (a) Topology. (b) Exemplar selection.}
	
	\label{toplogy and explar}
\end{figure}

\textit{\textbf{2) slArchive:}}\label{slArchive}  
In the initialization stage, we assign two particles with the same information—one from the \textit{non-G} subswarm and the other from the \textit{G} subswarm—to the same layer. Each layer has only one single-layer best vector, $\boldsymbol{L}$.  The two particles from \textit{n}th layer jointly update $\boldsymbol{L_n}$. Taking the minimization problem $f$ as an example, the update formula is shown in  Eq. (5).  This differs from the methods where each particle updates its own personal best vector \cite{hu2012adaptive,chen2018dynamic,yang2022adaptive}. In our architecture, this approach provides a uniform standard for utilizing the two abilities of the constructed vector. All single-layer best vectors form the \textit{slArchive}, and the correspondence of  relevant elements is illustrated in Fig. \ref{Correspondence}.

   \textit{\textbf{3) Two Update Channels:}}\label{Two Update Channels}  As we mentioned earlier,  the update mechanism in the \textit{global} PSO is good at aggregating particles, helpfully enhancing the exploitation ability. In particular, when adjusted with appropriate parameters, this mechanism can further enhance the exploitation ability in the later evolutionary process.  Here, we introduce two nearly heterogeneous update channels.   In our architecture, we refer to them as \textit{non-G-channel} and \textit{G-channel}, respectively. The \textit{non-G-channel} is a variant of the cognitive-only PSO, where the personal best vector is replaced by $\boldsymbol{Q_n}$ (It can come from any modified cognitive-only PSOs). The \textit{G-channel} is a variant of the \textit{global} PSO, where the personal best vector is replaced by the same $\boldsymbol{Q_n}$. The two channels are represented as follows:

\begin{equation} \label{non-G}
	\textit{non-G-channel}: 	 
	\begin{cases}
		\underset{non-G}{{v}_n^d} = w\cdot \underset{non-G}{{v}_n^d}+{c}\cdot {r} \cdot ({q_n^d}-\underset{non-G}{{x}_n^d})\\
		\underset{non-G}{{x}_n^d} = \underset{non-G}{{x}_n^d} + \underset{non-G}{{v}_n^d}
	\end{cases}
\end{equation}
and
\begin{equation} \label{G}
	\textit{G-channel}:
	\begin{cases}
		\underset{G}{{v}_n^d} = w\cdot \underset{G}{{v}_n^d}+{c_1}\cdot {r_1} \cdot ({q_n^d}-\underset{G}{{x}_n^d})\\+{c_2}\cdot {r_2} \cdot ({g}^d-\underset{G}{{x}_n^d})\\
		\underset{G}{{x}_n^d}=\underset{G}{{x}_n^d}+\underset{G}{{v}_n^d}
	\end{cases},
\end{equation}
 where  $\boldsymbol{Q_n} = (q^1_n,q^2_n,...,q^D_n )$. Clearly, our proposed two update channels contain the same constructed vector. This operation allows the two abilities of one  constructed vector to be aligned with the properties of their respective channels. 
	
	 \textit{\textbf{4) Information Pool and Operator:}}\label{Information Pool and operator}  The information pool and operator are conceptual components, i.e., replaceable components. These two components are closely linked. In a specific modified cognitive-only PSO, the information pool stores relevant information, and the operator constructs vectors using specific methods. For example, in cognitive-only PSO, the information pool consists of all personal best vectors, and the operator selects the \textit{n}th personal best vector for the \textit{n}th particle. In CLPSO, the information pool similarly consists of all personal best vectors, and the operator is CL strategy to construct distinct vectors for particles. However, unlike these PSOs where one particle acts as a unit \cite{kiran2017particle, moazen2023pso}, our architecture is layer-based. Thus, it is necessary to redefine some basic terms to facilitate information storage and vector construction, thereby helping embed the vector-construction methods from modified cognitive-only PSOs  into our proposed architecture.

	\begin{itemize}
		\item [$\bullet$]\textbf{Single-layer best vector and single-layer best position.} The two proposed terms correspond to the personal best vector and personal best position in the traditional PSOs.  In our architecture, each layer has a single-layer best vector and a corresponding best position, which are jointly updated by the two particles in the same layer.
	\end{itemize}

		\begin{itemize}
		\item [$\bullet$]\textbf{Topology.} In traditional PSOs, every particle participates in each iteration, and all particles are included in the topology. However, in our architecture, only one particle within each layer is allowed to participate in each iteration. It means only half of the particles are active in each iteration. Therefore, we suggest that only these active particles should be included in the topology in each iteration. For example, Fig. \ref{toplogy and explar}(a) illustrates the particles included in the \textit{ring} topology, when the number of layers is five.
	\end{itemize}

		\begin{itemize}
	\item [$\bullet$] \textbf{Exemplar selection.} In traditional PSOs, all particles are considered to be  selected based on specific rules. In our architecture,   only  half of  particles are  active in each iteration. Thus,  we suggest that only these active particles should be considered for selection. Fig. \ref{toplogy and explar}(b) illustrates how the two nearest particles are selected as exemplars for the first layer when the number of layers is five.
\end{itemize}

	With the help of these basic terms, we can embed a series of vector-construction methods from modified cognitive-only PSOs  into our  architecture.  This represents the generality of our architecture.  Section \ref{Experimental Setup} gives  the details of how the vector-construction methods from cognitive-only PSO and CLPSO are embedded into our architecture.
	
			\subsection{ABS Strategy}\label{Adaptive balance Search Strategy}	
	 The ABS Strategy consists of three components: the  cap limiter, the  R \& P box,  and the adaptive selector. These components are described in detail below:

			\textit{\textbf{1) Cap Limiter:}}\label{Cap Limiter}   The cap limiter first needs a total upper threshold, $M$,  for the total number of utilization of the constructed vectors, which is a common and necessary operation \cite{liang2006comprehensive,zhan2009orthogonal}.  This limits the number of times that a constructed vector can exist in the evolutionary process. Without this setting,   particles are likely to keep flying under the influence of fixed vectors, making it difficult to explore other directions. In addition, to avoid an equal emphasis on the utilization of  two abilities throughout the evolutionary process, we decompose the total upper threshold into two upper thresholds, named exploration upper threshold $\underset{non-G}  {M}$ and exploitation upper threshold $\underset{G}  {M}$. The exploration  upper threshold and exploitation upper threshold limit the number of utilization of the constructed vectors to guide particle exploration and exploitation, respectively.  Both upper thresholds are regulated by the number of function evaluations, calculated by
			
				\setcounter{equation}{7}
			\begin{equation} \label{dynamic adjustment function m1}
				\begin{cases}
					\underset{non-G}  {M}=\lceil  {M} \cdot(1-\dfrac{FEs}{FEs_{max}}) \rceil \\
					\underset{G}  {M}=\lfloor  {M} \cdot(\dfrac{FEs}{FEs_{max}}) \rfloor \\
				\end{cases},
			\end{equation}	where $FEs_{max}$ denotes the maximum number of function evaluations, $FEs$ is the function evaluations counter, $\lceil\cdot\rceil$ represents rounding up, $\lfloor\cdot\rfloor$ represents rounding down. Rounding up helps emphasise the ability of constructed vectors to guide particle exploration and mitigate the decline of diversity in the early evolutionary process, which shown in detail in Section \ref{General Algorithm of CHCLPSO-ABS}. Moreover, when $M$ is fixed, $\underset{non-G} {M}$ decreases while $\underset{G} {M}$ increases, as $FEs$ increases. Such a trend exhibits the emphasis on the ability of constructed vectors to guide particle exploration in the early evolutionary process, and the emphasis on the ability of constructed vectors to guide particle exploitation in the later. Overall, the adaptive setting of the exploration and exploitation upper thresholds helps the  swarm emphasise  more exploration in the early evolutionary process while more exploitation in the later. The investigation of the total upper threshold is presented  in Section \ref{Analysis of the Sensitivity of parameter}.

		\textit{\textbf{2) R $\&$ P Box:}}\label{R P Box:}	What we cannot ignore is that one particle may or may not find a better solution, driven by one ability of the constructed vector. Therefore, if one ability of the constructed vector is able to guide the particle to find a better solution, accordingly, we should reward its existence time to drive the particle to find a better solution than before. Otherwise, we should penalize its existence time to avoid the waste of computational resources. For such purpose, we build an R \& P box, which is used to determine whether to reward or penalize the number of times that the two abilities of each constructed vector are utilized. Here, we introduce three parameters: $	\underset{non-G} {\alpha_n}$, $\underset{G} {\alpha_n}$, and $ \beta_n$. The first parameter tracks the times that the ability of $n$th constructed vector to guide particle exploration has been utilized. The second 	indicates the times that the ability of $n$th constructed vector to guide particle exploitation has been utilized. The third counts the times that $\boldsymbol{L}_n$ is updated by \textit{non-G} particles under a fixed $\boldsymbol{Q}_n$. All three parameters are reset to $0$ when the vector is reconstructed. When one constructed vector is prepared to be utilized, the two rules,  controlled by the updated states of single-layer best vectors, are suggested to be followed. Rule 1 is used for regulating the number of times that the ability of each constructed vector to guide particle exploration is utilized. Rule 2 is used for regulating the number of times that the ability of each constructed vector to guide particle exploitation is utilized. 
		
				\begin{itemize}
			   \item [$\bullet$] Rule 1 (exploration): 
			   
			   If $\boldsymbol{L}_n$ is updated, it indicates that the exploration ability is valuable. Reset $\underset{non-G} {\alpha_n} = 0$  to reward the number of  utilization, and set $ \beta_n = \beta_n +1 $. If $\boldsymbol{L}_n$ is not updated, it suggests the ability is less valuable. Set $ \underset{non-G} {\alpha_n} = \underset{non-G} {\alpha_n} +1 $ to reduce the number of  utilization, and set $ \beta_n = \beta_n $.
						\end{itemize}
					
							\begin{itemize}	
		\item [$\bullet$] 	Rule 2 (exploitation): 
		
		If $\boldsymbol{G}$ is updated, it indicates that the exploitation ability is valuable.   Reset $\underset{G} {\alpha_n} = 0$ to reward the number of  utilization. If $\boldsymbol{L}_n$ is updated but $\boldsymbol{G}$ is not, this indicates that the ability has a slight value. Perform $\underset{G}  {\alpha_n} =\underset{G}  {\alpha_n}$ to reward the number of  utilization once more. If $\boldsymbol{L}_n $ is not updated,  execute $\underset{G}  {\alpha_n} = \underset{G}  {\alpha_n}+1 $ to reduce the number of  utilization. 
						\end{itemize}
					
					By applying the above  reward and punishment rules, our architecture can then accurately calculate the number of times that the two abilities of the constructed vectors are utilized, respectively.

		\begin{algorithm}[!t]
			\renewcommand{\algorithmicrequire}{\textbf{Input:}}
			\renewcommand{\algorithmicensure}{\textbf{Output:}}
			\caption{\textbf{:} Pseudocode of ABS strategy for \textit{n}th layer}
			\label{strategy 2}
			\begin{algorithmic}[1]
				\Require  $\boldsymbol{L}_n $, Information pool, $\underset{non-G}{\alpha_n}$, $\underset{G} {\alpha_n}$, $\beta_n$, $M$, $\underset{non-G}{\boldsymbol{X}_n}$, $\underset{non-G}{\boldsymbol{V}_n}$, $\underset{G}{\boldsymbol{X}_n}$, $\underset{G}{\boldsymbol{V}_n}$, $FEs$, $FEs_{max}$; 
				\Ensure  $\underset{non-G}{\alpha_n}$, $\underset{G} {\alpha_n}$, $\beta_n$, $\boldsymbol{L}_n$, $\boldsymbol{G}$, $FEs$; 		
				
				\State Calculate $\underset{non-G} {M}$ and  $\underset{G} {M}$ according to (\ref{dynamic adjustment function m1});
				\If{{$(\beta_n \neq 0  \&\&  \underset{non-G} {\alpha_n} > \underset{non-G} {M})  \|   \underset{G} {\alpha_n} > \underset{G} {M}$}}
				\State$\underset{non-G} {\alpha_n}=0$, $\underset{G} {\alpha_n}=0$,  $\beta_n =0$,  and reconstruct $\textbf{\textit{Q}}_{n}$ according to information pool and operator;
				\EndIf
				
				\If {$\underset{non-G} {\alpha_n}  \leq  \underset{non-G} {M}$}
				\State Update $\underset{non-G}{\boldsymbol{V}_n}$ , $\underset{non-G}{\boldsymbol{X}_n}$ by (\ref{non-G}), $FEs=FEs+1$;
				
				\If {$f(\underset{non-G}{\boldsymbol{X}_n}) \geq f(\boldsymbol{L}_n)$}
				\State $\underset{non-G} {\alpha_n} =\underset{non-G} {\alpha_n}+1$,  $\beta_n=\beta_n$;
				\Else
				\State $\underset{non-G} {\alpha_n}=0$, $\beta_n=\beta_n+1$, and update $\boldsymbol{L}_n$;
				\If{$f(\boldsymbol{L}_n) < f(\boldsymbol{G})$}
				\State Update $\boldsymbol{G}$;
				\EndIf
				\EndIf
				\ElsIf{$  (\underset{non-G} {\alpha_n} >  \underset{non-G} {M}  \&\&  \beta_n=0)  \&\&  \underset{G} {\alpha_n} \leq \underset{G} {M} $}
				
				\State Update $\underset{G}{\boldsymbol{V}_n}$, $\underset{G}{\boldsymbol{X}_n}$ by (\ref{G}), $FEs=FEs+1$;
				\If {$f(\underset{G}{\boldsymbol{X}_n}) \geq f(\boldsymbol{L}_n)$}
				\State $\underset{G} {\alpha_n} =\underset{G} {\alpha_n}+1$;
				\Else                     
				\State $\underset{G} {\alpha_n} =\underset{G} {\alpha_n}$, and update $\boldsymbol{L}_n$;
				\If{$f(\boldsymbol{L}_n) < f(\boldsymbol{G})$}
				\State $\underset{G} {\alpha_n}=0$, and update $\boldsymbol{G}$;
				\EndIf
				\EndIf							     
				\EndIf
				
				\State Update information pool;
				\State \Return   $\underset{non-G} {\alpha_n}$, $\underset{G} {\alpha_n}$, $\beta_n$, $\boldsymbol{L}_n$, $\boldsymbol{G}$, information pool, and $FEs$.
			\end{algorithmic}
		\end{algorithm}

			\begin{algorithm}[!t]
			\renewcommand{\algorithmicrequire}{\textbf{Input:}}
			\renewcommand{\algorithmicensure}{\textbf{Output:}}
			\caption{\textbf{:} Pseudocode of CH\textit{x}PSO-ABS architecture}
			\label{VRD-PSO}
			\begin{algorithmic}[1]
				\Require Simple objective optimization problem with $f$; search space $S^D$, $D$ is the number of decision variables; the population size $N$; the maximal number of function evaluations $FEs_{max}$; the total upper threshold $M$; 
				\Ensure  $\boldsymbol{G}$; 	\\
				/*  Initialization */
				\State 	$k_{max}=\lfloor FEs_{max}/N \rfloor$,  $k=0$,	$FEs=0$; 
				
				\For{1 $\leq$ \textit{n} $\leq$ \textit{N}}
				\State Randomly initialize $\boldsymbol{X}_n$ and $\boldsymbol{V}_n$, and evaluate $f(\boldsymbol{X}_n)$;
				\State Set $\boldsymbol{L}_n=\boldsymbol{X}_n$, and update $\boldsymbol{G}$;
				\EndFor
				\State Store key information into information pool, and construct $\boldsymbol{Q_n}, n = 1,...,N$ for all layers according to specific operator.

				\State $k=1$, $FEs=N$;
				\State Set that $\underset{non-G}{\boldsymbol{X}_n}$ and $\underset{G}{\boldsymbol{X}_n} $ are equal to $\boldsymbol{X}_n$, that $ \underset{non-G}{\boldsymbol{V}_n}$ and $\underset{G}{\boldsymbol{V}_n}$ are equal to $\boldsymbol{V}_n$, that $\underset{non-G} {\alpha_n}$, $\underset{G} {\alpha_n}$, $\beta_n$ are equal to 0; \\
				
				/* Main Loop */
				\While{$k \leq k_{max} \&\& FEs \leq  FEs_{max}$}
				\State $k=k+1$;
				\For{1 $\leq$ \textit{n} $\leq$ \textit{N}}
				\State Update $\underset{non-G} {\alpha_n}$, $\underset{G} {\alpha_n}$, $\beta_n$, $\boldsymbol{L}_n$, $\boldsymbol{G}$, information pool, and $FEs$ by Algorithm \ref{strategy 2};
				\EndFor
				\EndWhile
				\State \Return  $\boldsymbol{G}$.
			\end{algorithmic}
		\end{algorithm}

			 	\textit{\textbf{3) Adaptive Selector:}}\label{Adaptive Selector} In addition to the above set of operations, another important operation is needed to   designate  which ability of the constructed vectors should be utilized. For such purpose, we construct an adaptive selector based on the cap limiter and R $\&$ P box. The adaptive selector includes three conditions, shown as follows:	
			 				\begin{itemize}
			 	
			 	\item [$\bullet$] Condition 1 (exploration): 
			 	
			 	$\underset{non-G} {\alpha_n}  \leq  \underset{non-G} {M}$.
			 	
			 \end{itemize}
		 \begin{itemize}
			 			 	\item [$\bullet$]Condition 2 (exploitation): 
			 			 	
			 			 	$  (\underset{non-G} {\alpha_n} >  \underset{non-G} {M}  \& \&  \beta_n=0)  \&\&  \underset{G} {\alpha_n} \leq \underset{G} {M} $.
			 
		 \end{itemize}
	 \begin{itemize}
	 			 	\item [$\bullet$]  Condition 3 (reconstruction): 
	 			 	
	 			 	$(\beta_n \neq 0 \&\& \underset{non-G} {\alpha_n} > \underset{non-G} {M}) \|  \underset{G} {\alpha_n} > \underset{G} {M}$.
	 
 \end{itemize}

 When condition 1 is satisfied, it indicates that we should utilize the ability of the constructed vector to guide particle exploration.  When condition 2 is satisfied, it indicates that we should utilize the ability of the constructed vector to guide particle exploitation. In general, one fruitful algorithm typically emphasises exploration in the early evolutionary process, while  exploitation in the later. Therefore, it is logical to prioritize the ability of constructed vectors to guide particle exploration, i.e., condition 1 has priority over condition 2. In addition, to prevent the over-reliance on the ability of $n$th constructed vector to guide particle exploitation,  we set that only when condition 1 is not met and $\boldsymbol{L}_n$ is not updated under condition 1, is the ability of the constructed vector to guide particle exploitation  utilized. In addition, when condition 3 is triggered, we need to reconstruct the vector. At the same time, reset the relevant parameters to $0$ so that condition 1 is automatically satisfied. This can guarantee to seamlessly utilize the ability of the reconstructed vectors to guide particle exploration. 
 
One undeniable point is that the utilization of two abilities of constructed vectors is operated by particles. In our architecture, \textit{non-G} and \textit{ G} particles, two channels, and the two abilities of constructed vectors are regularly one-to-one, i.e., \textit{non-G} particles work in the \textit{non-G-channel} throughout the evolutionary process, and the ability of the constructed vector to guide particle exploration is mainly utilized by \textit{non-G-channel}, thereby this ability is mainly operated by \textit{non-G} particles; similarly, so do the \textit{G} particles. Thus, conditions 1 and 2 also implicitly select the particles to participate in the evolution. Specifically,  \textit{non-G} particles are employed into \textit{non-G-channel} to prefer exploring under condition 1, and \textit{G} particles are employed into  \textit{G-channel} to prefer exploiting under condition 2. Therefore, the adaptive selector also implicitly controls the proportion of particles in the two channels involved in evolution.  After all, precise control of the proportion of particles with different behaviors can help improve the performance of algorithms \cite{de2009heterogeneous}.

  In summary, the combined operation of these components enables the ABS strategy to control the proportion of particles in the two channels involved in evolution, thereby determining when vectors should be reconstructed and which ability of constructed vector is utilized. The pseudocode for the ABS strategy is presented in Algorithm \ref{strategy 2}.

	\subsection{General Algorithm of CH\textit{x}PSO-ABS Architecture}\label{General Algorithm of CHCLPSO-ABS}

	\begin{figure}[!t]
		\centering
		\includegraphics[width=2.6in]{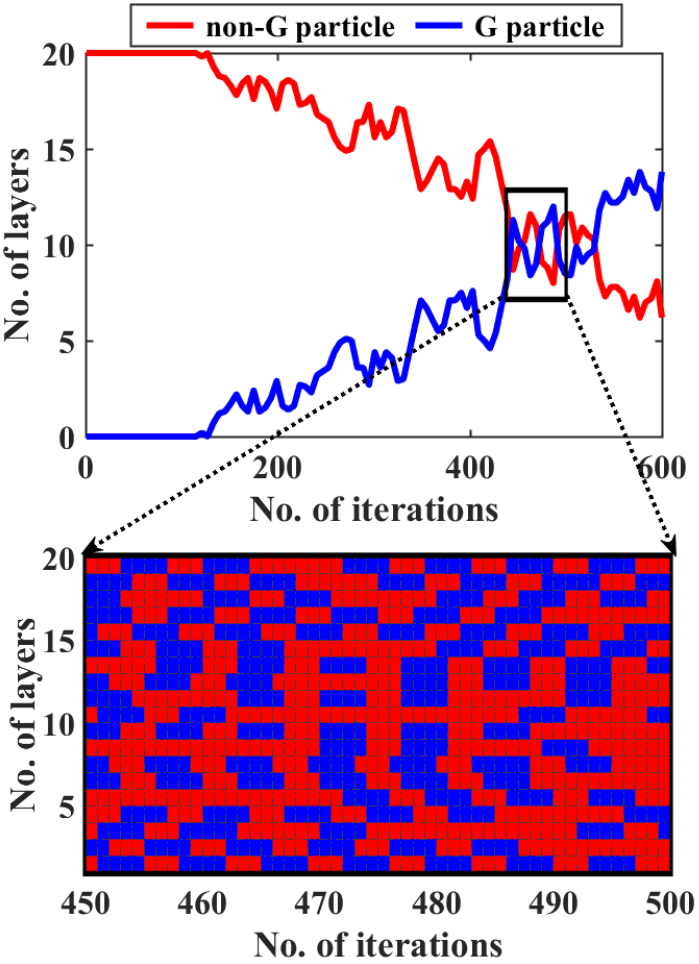}
		\caption{The number of employed particles during the evolutionary process number of iterations (up). The combination of particles employed  on different iterations (down).}
		\label{Plot of changes in sub-swarms participation}
	\end{figure}	
	 The pseudocode of CH\textit{x}PSO-ABS architecture is presented in Algorithm \ref{VRD-PSO}. 
	 
\begin{figure} [!]
	 	\subfigure[]{
	 		\includegraphics[width=1.7in]{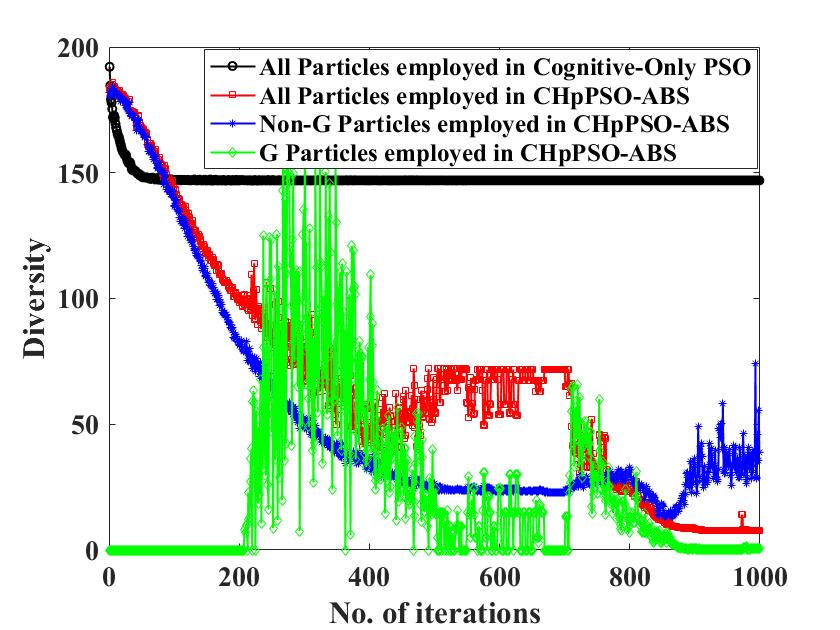}\\
	 	}%
	 	\hspace{-2mm}
	 	\subfigure[]{
	 		\includegraphics[width=1.7in]{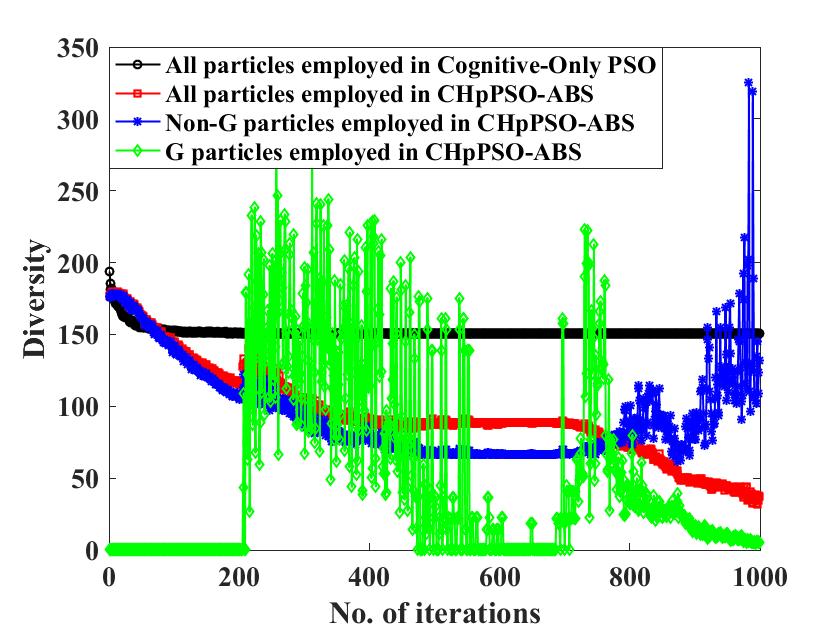}\\
	 	}
	 	\subfigure[]{
	 		
	 		\includegraphics[width=1.7in]{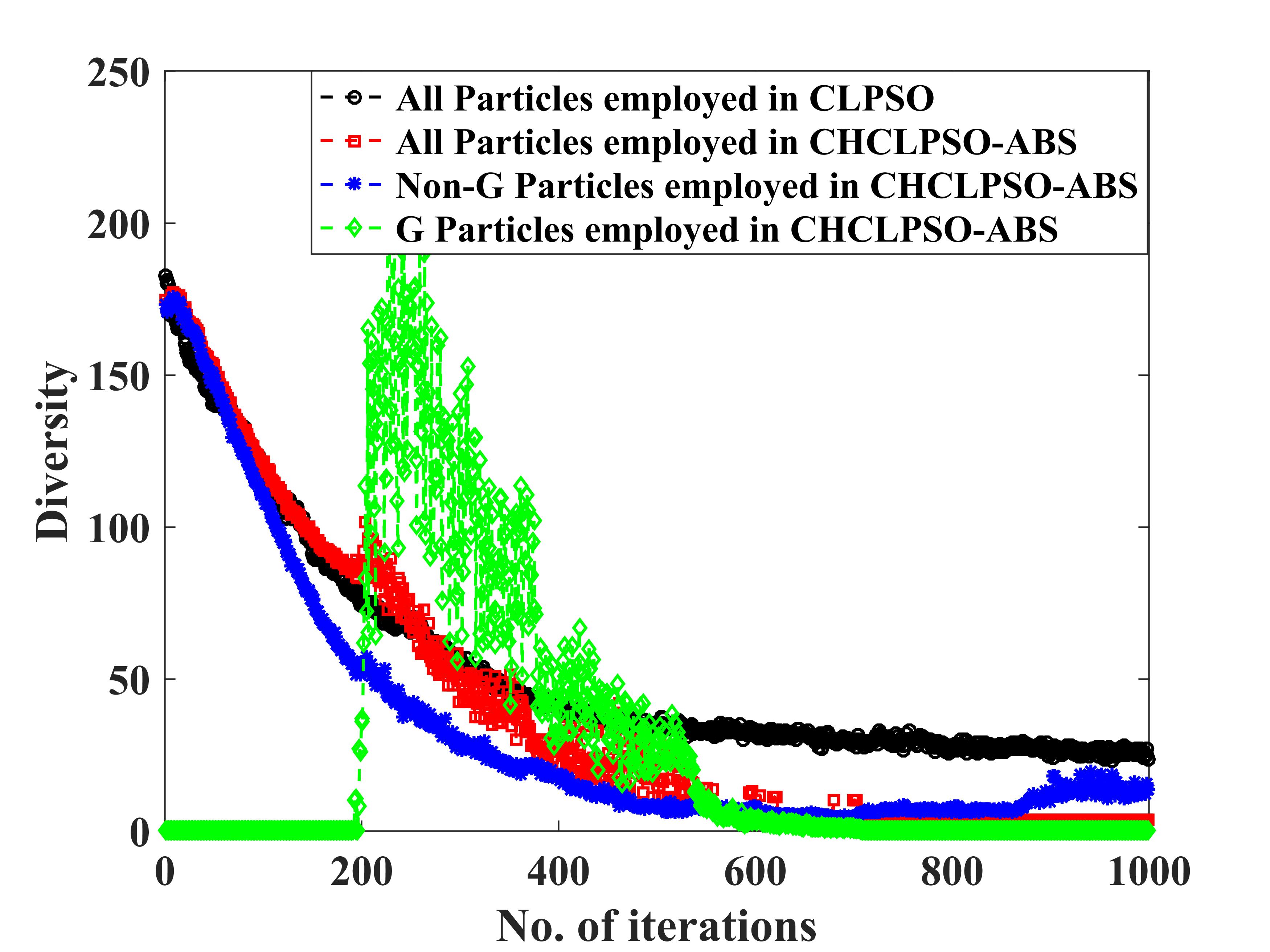}\\
	 	}%
	 	\hspace{-2mm}
	 	\subfigure[]{
	 		\includegraphics[width=1.7in]{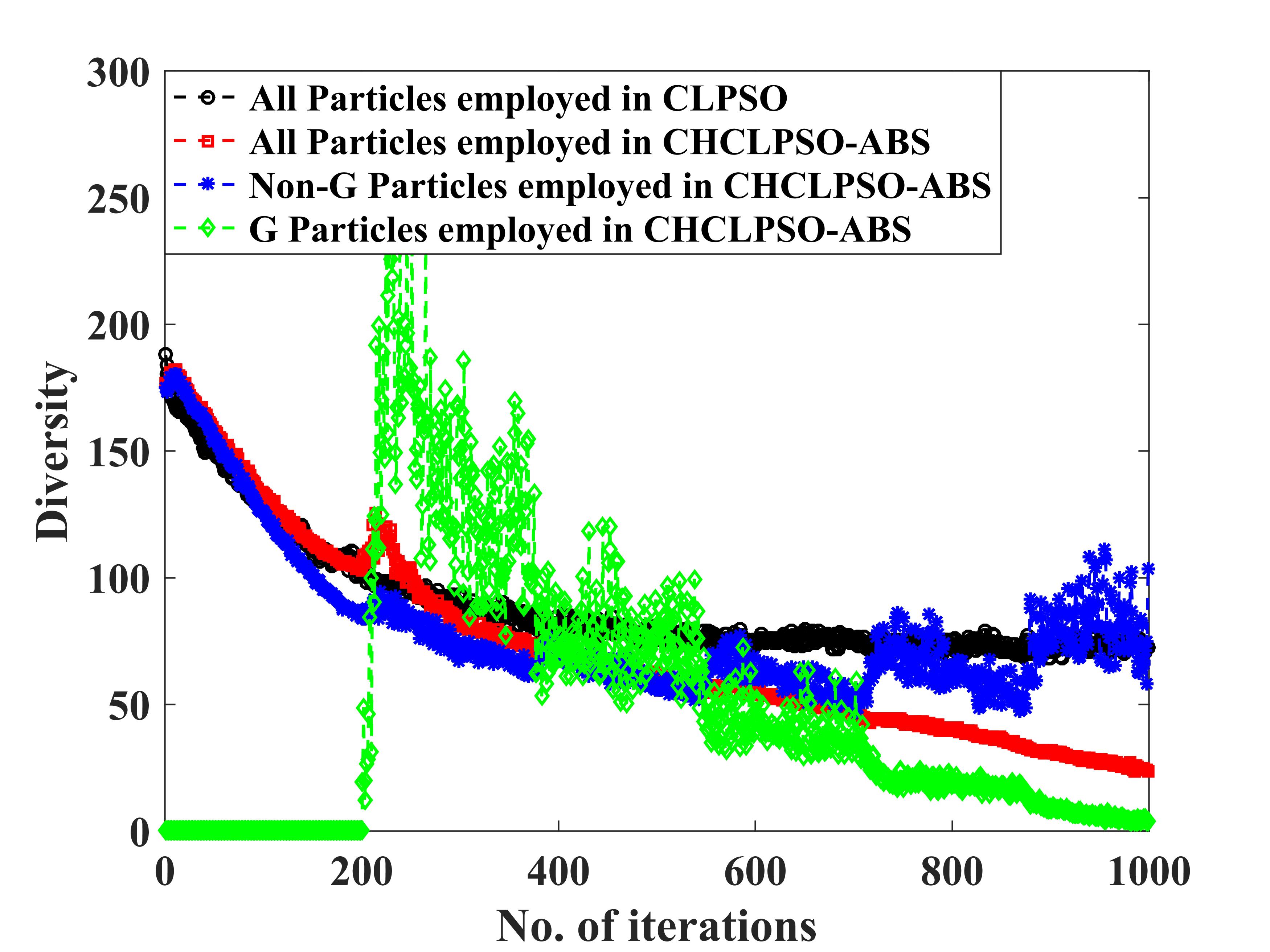}\\
	 	}
	 	\caption{
	 		Trend plots of swarm diversity indicator. (a) Cognitive-only PSO and CHpPSO-ABS on the Sphere function. (b) Cognitive-only PSO and CHpPSO-ABS on the Rastrigin function. (c) CLPSO and CHCLPSO-ABS on the Sphere function. (d) CLPSO and CHCLPSO-ABS on Rastrigin functions.}
	 	
	 	\label{Trend plots of swarm diversity indicator.}
	 \end{figure}

	   In each iteration, only $N$ particles are selected to participate in the evolution due to the adaptive selector, though a total of $2N$ particles are available. This results in $2^N$ possible combinations of particles in each iteration. Fig. \ref{Plot of changes in sub-swarms participation} illustrates this unique combination with the case of $N = 20$. The upper subplot shows the trend of the number of \textit{Non-G} and \textit{G} particles employed during the evolutionary process. We can observe that the number of employed \textit{non-G} particles decreases, while the number of employed \textit{G} particles increases as the evolutionary process proceeds. Intuitively, the proportion of particles in the two channels is precisely regulated. Essentially, this shows that our architecture emphasises the ability of constructed vectors to guide particle exploration in the early evolutionary process, while the ability of constructed vectors to guide particle exploitation in the later. Notably, the \textit{G} subswarm is not employed until around the 100th iteration, further mitigating the loss of diversity in the early evolutionary process, as expected from Eq. (\ref{dynamic adjustment function m1}). These results show that our architecture can avoid premature convergence and improve convergence accuracy. Moreover, the lower subplot shows that the combinations of \textit{non-G} and \textit{G} particles vary at different iterations. The alternating red and blue blocks illustrate the flexible combinations of particles, the precise control of the particles involved in evolution in two channels, as well as the flexible utilization of the constructed vectors. Furthermore, half of the particles are active while half remain still during the evolutionary process, implicitly causing subtle diversity jumps, which may enhance the swarm activity. This phenomenon is visualized in Fig. \ref{Trend plots of swarm diversity indicator.}.

	\section{Experiments Evaluation and  Discussion}\label{experiment}
In this section, a series of experiments are designed. Firstly, the  generalization performance of our architecture is verified. Specifically, we analyse  the exploration and exploitation performance, its convergence accuracy, and the sensitivity to the total upper threshold. Secondly,  we compare CHCLPSO-ABS (one of our proposed algorithms, which is a combination of CLPSO and our architecture) with seven popular algorithms to demonstrate its convergence performance. Finally, an experiment is designed to test the stability of CHCLPSO-ABS under different values of $FEs_{max}$.

	\subsection{ Experimental Setup} \label{Experimental Setup}
	\textit{1) Benchmark Functions:} The CEC 2013 \cite{liang2013problem} and CEC 2017 \cite{wu2017problem} test suites  are  widely  used to  evaluate the performance of algorithms.  In particular, the 30 test benchmark functions in the CEC 2017 test suite with boundary constraints are all complex, which are  often used to comprehensively assess algorithmic performance. However,  an official statement acknowledges that the Shifted and Rotated Sum of Different Power Function in the CEC 2017 test suite exhibits instability, posing challenges in correctly evaluating algorithmic efficacy. Therefore, the remaining 57 functions are involved in our experiments, which are divided into three groups: Unimodal Functions, Simple Multimodal Functions, and Complex Multimodal Functions, as shown in Table S-I in the attached \textbf{Supplementary Material}. And  the optimal value of each function is labeled as  $F^*_i$, where $i$ represents the number of the test function. In all our subsequent experiments, the error $F_i-F^*_i$ is used to calculate the performance of the algorithms on each test function, where $F_i$ denotes the experimental value on  $i$th test function.

		\textit{2) CHpPSO-ABS and CHCLPSO-ABS Algorithms:} Among the modified cognitive-only PSO variants, we choose two typical algorithms: cognitive-only PSO and CLPSO. In cognitive-only PSO, the vectors used for guiding each particle, i.e., the personal best vectors, are constructed from their own personal best information. As a result, the constructed vectors in this algorithm only include the information of their corresponding particles. By contrast, the vectors used to guide each particle in CLPSO include the information of the whole swarm. In terms of the information components of the constructed vectors, they somewhat represent the middle-low and middle-high levels, respectively. Thus, we combine these two algorithms with our proposed architecture to demonstrate the  generalization performance of our proposed architecture. CHpPSO-ABS is a combination of the cognitive-only PSO and our architecture. Based on the basic terms defined in Section \ref{Information Pool and operator}, we only need to set the information pool to contain single-layer best vectors. We set the operator to only select the corresponding single-layer best vector for each layer. Then CHpPSO-ABS is obtained. CHCLPSO-ABS is a combination of CLPSO and our architecture. Similarly, we set the information pool to contain single-layer best vectors and set the operator as the CL strategy, then the CHCLPSO-ABS is obtained. Not come singly but in pairs, a series of other modified cognitive-only PSOs can be also embedded into our architecture. The pseudocode of the two algorithms is presented in the attached \textbf{Supplementary Material}.
		
	\textit{3) Parameter Settings of PSOs:} Eight state-of-the-art peer algorithms are  simultaneously compared with our proposed algorithm: Cognitive-only PSO \cite{engelbrecht2010heterogeneous}, HPSO-TVAC \cite{ratnaweera2004self}, CLPSO \cite{liang2006comprehensive}, OLPSO \cite{zhan2009orthogonal}, HCLPSO \cite{lynn2015heterogeneous}, EPSO \cite{lynn2017ensemble}, AWPSO \cite{liu2019novel}, and MAPSO \cite{wei2020multiple}. The parameter configurations of these algorithms are taken from their corresponding papers.  All the algorithms are  executed on a PC with an Intel(R) Core(TM) i5-10400 CPU at 2.90 GHz with 64 GB of RAM. The parameter configurations of  these algorithms as well as CHpPSO-ABS and CHCLPSO-ABS are presented in Table S-II in the attached \textbf{Supplementary Material}.

	\begin{table}[!t]
		\centering
		\caption{Comparison Results  on CEC 2013 Test Suite Among Cognitive-Only PSO, CLPSO, CHpPSO-ABS, and CHCLPSO-ABS}
		\includegraphics[width=3.4in]{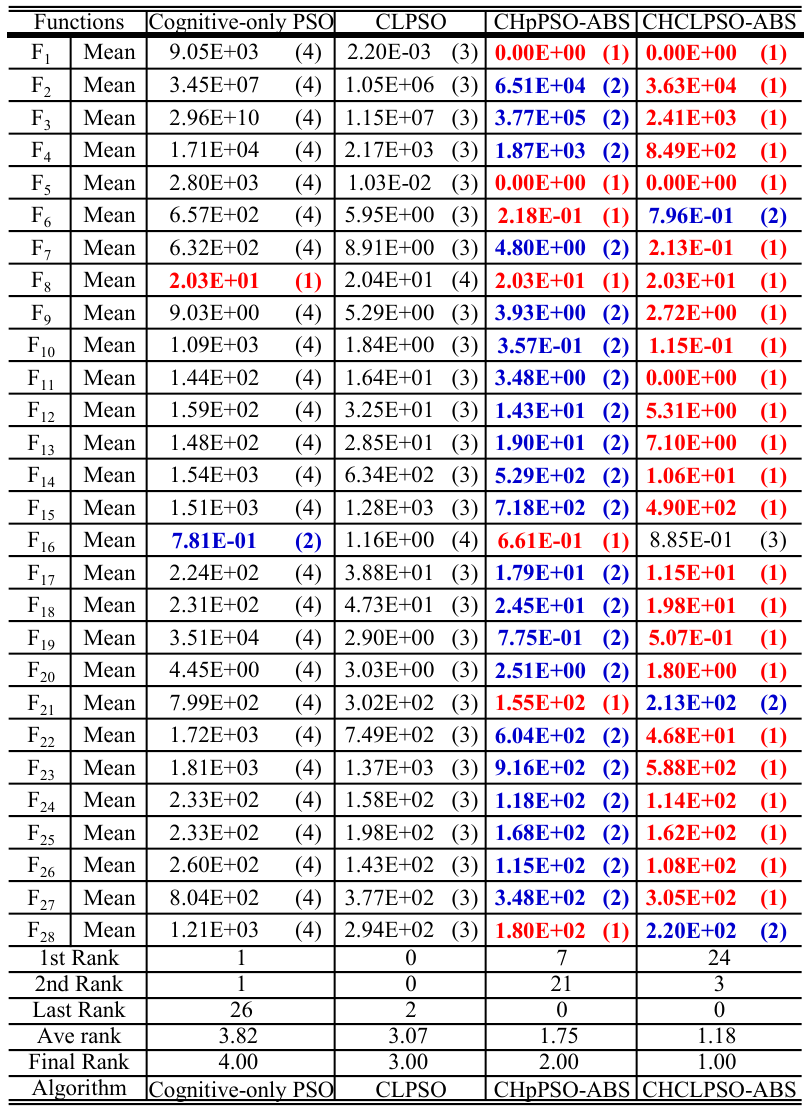}
		
		\label{comparison results  on CEC 2013 test functions among Cognitive-Only PSO, CLPSO, CHpPSO-ABS and CHCLPSO-ABS}
	\end{table}
	
		\textit{4) Swarm Diversity Indicator:}	\label{Indicator}
	The indicator, swarm diversity, is widely used to identify whether the swarm  conducts exploration or exploitation \cite{olorunda2008measuring}. The diversity is commonly measured by 
	\begin{equation} \label{diversity eqation 1}
		\begin{cases}
			Divesity(k)=\dfrac{\sum_{n=1}^{N} \sqrt{\sum_{d=1}^{D} \big(x_i^d(k)-\bar{x}^d(k)\big)^2}}{N}  \\
			\bar{x}^d(k)=\dfrac{\sum_{n=1}^{N} x_i^d(k)}{N}
		\end{cases},
	\end{equation}
where $k$ indicates the number of iterations, $x_i$ represents the position of $i$th employed particle, $\bar{x}$ denotes the average position of all employed particles, and $N$ denotes the total number of employed particles.

	\textit{5) Wilcoxon Signed-Rank Test:} As  PSO is one of the nondeterministic algorithms, a statistical hypothesis test is applied to assess the significance of the differences in results. A widely used nonparametric statistical hypothesis test, the  Wilcoxon Signed-Rank Test, is employed at a significance level of $\alpha = 0.05$. In this paper, the symbols (+) / (-) indicate the proposed algorithms perform significantly better / worse than the compared algorithms, respectively,  the symbol (=) denotes that there is no significant difference  between the proposed algorithms and the compared algorithms.

	\subsection{Experimental Results on  Generalization Performance of  CHxPSO-ABS Architecture} \label{Experimental Results on  Generalizability Performance of  CHxPSO-ABS}
Our architecture can embed cognitive-only PSO (one basic algorithm) and a series of modified cognitive-only PSOs. In order to verify the generalization performance of our architecture, we embed cognitive-only PSO and a typical modified cognitive-only PSO, CLPSO, into our architecture, further obtaining two canonical algorithms, CHpPSO-ABS (It can be viewed as one basic algorithm deriving from our architecture) and CHCLPSO-ABS, respectively. As analysed in Subsection \ref{Experimental Setup}, they somewhat represent the PSO variants with lower or higher performance deriving from our architecture, respectively. Therefore, in this section, we rely on the two canonical algorithms to analyse the generalization performance of the CH\textit{x}PSO-ABS architecture, specifically focusing on its exploration and exploitation performance, convergence accuracy, and parameter sensitivity (robustness).
	
	\textit{1) Analysis of Exploration and Exploitation Performance:}  \label{Analysis of Exploration and Exploitation Performance}
	 To validate the advantages of our architecture, we calculate the diversity  of cognitive-only PSO, CHpPSO-ABS, CLPSO, and CHCLPSO-ASB  on the 10-D Sphere Function (a unimodal function) and Rastrigin Function (a multimodal function).  Fig. \ref{Trend plots of swarm diversity indicator.}  shows the diversity induced by all employed particles in 	different algorithms on the two functions. 
	 
	In Fig. \ref{Trend plots of swarm diversity indicator.}(a),  we find that the diversity of cognitive-only PSO alone decreases to a certain level and then stabilizes. This is expected since the particles tend to make only slight local movements.  This phenomenon although apparently shows that these particles will be far away from each other, virtually shows the particles have been confined to small spaces  after a certain number of iterations.  In contrast, with the help of our architecture, the diversity of all employed particles in CHpPSO-ABS shows a trend of overall decrease. This is mainly attributed  to  the effective  utilization of the constructed vectors by  employing different particles during the evolutionary process. At the same time, the curve of CHpPSO-ABS also features some jumps, which is consistent with our inferences in Fig. \ref{Plot of changes in sub-swarms participation}. Moreover, we observe that the \textit{non-G} particles are employed, while the \textit{G} particles are not, i.e., \textit{G} particles remain still, before 200 iterations due to the cap limiter. This is a good phenomenon as it ensures strong exploration in the early evolutionary process. Around 200th iterations, \textit{G} particles are involved in evolution and exhibit greater diversity, warming up the diversity.  As the evolutionary process proceeds, the diversity of \textit{G} particles overall decreases and is even smaller than \textit{non-G} particles in the later evolutionary process. This indicates our architecture exhibits strong exploitation with the help of the \textit{G-channel} in the later evolutionary process. Besides, the diversity of \textit{non-G} particles picks up slightly in the later evolutionary process. This shows that our architecture may meanwhile reduce the risk of falling into local optimums with the help of \textit{non-G-channel} in the later evolutionary process. 
	
	In Fig. \ref{Trend plots of swarm diversity indicator.}(b), the diversity of \textit{non-G} particles in CHpPSO-ABS increases greater in the later evolutionary stages compared to Fig. \ref{Trend plots of swarm diversity indicator.}(a), even surpassing that of cognitive-only PSO. This is a positive outcome. Because greater diversity means stronger resistance to the risk of falling into a local optimum posed by multimodal problems. Therefore, our architecture has a strong advantage in dealing with multimodal problems. These positive effects are largely attributed to the special designs in our proposed architecture.
	 
Similar results can be observed in Fig. \ref{Trend plots of swarm diversity indicator.}(c) and Fig. \ref{Trend plots of swarm diversity indicator.}(d). Since the analysis process is essentially the same, we will not dwell further. However, some differences are noticeable in Fig. \ref{Trend plots of swarm diversity indicator.}(a) and Fig. \ref{Trend plots of swarm diversity indicator.}(c). One marked difference is that the declining trend of diversity in CLPSO is very different from cognitive-only PSO. Another is the jumps of diversity in CHCHPSO-ABS are relatively milder compared to CHpPSO-ABS. These differences can be explained by the fact that constructed vectors in the CL strategy contain additional information. This allows for better communication among particles, thereby enhancing the abilities of the constructed vectors to guide particle exploration and exploitation. After all, effective communication among particles is crucial for an algorithm to perform well \cite{blackwell2018impact}. Similar differences can also be concluded in Fig. \ref{Trend plots of swarm diversity indicator.}(b) and Fig. \ref{Trend plots of swarm diversity indicator.}(d).

	The above results highlight the effectiveness of our proposed architecture for these modified cognitive-only PSO variants. This is further demonstrated in the next subsection through convergence experiments. Additionally, the results confirm the importance of constructing proper vectors and ensuring effective communication among particles, although this is not the main focus of our paper.

		\begin{table}[!t]
		\centering
		\caption{ Mean Results of CHpPSO-ABS and CHCLPSO-ABS at Different $M$}
		\includegraphics[width=3.5in]{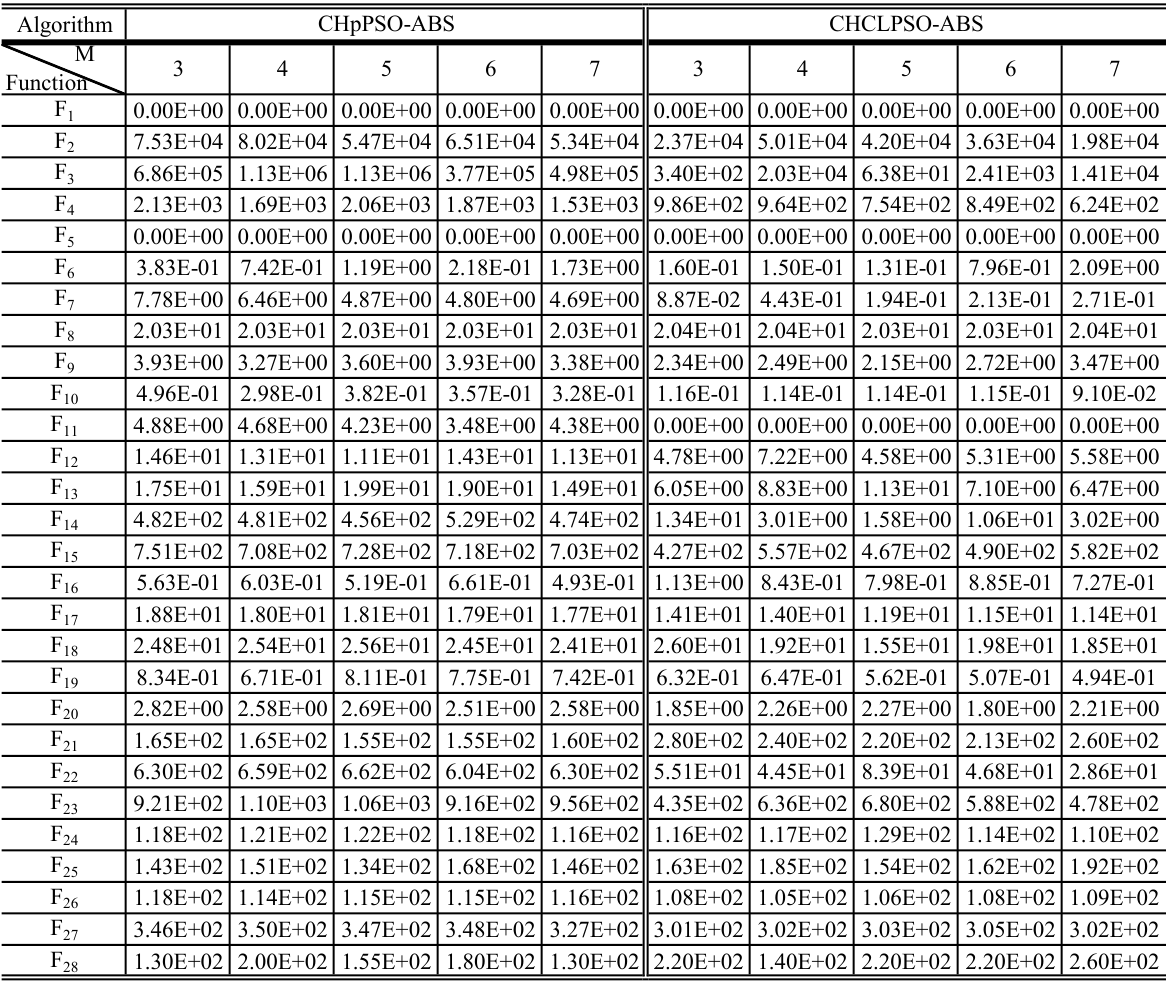}
		
		\label{Mean of CHxPSO-ABS Algorithms}
	\end{table}

	\textit{2) Analysis of Convergence Accuracy:}  \label{Analysis of Convergence Accuracy} 	In this subsection, we examine the convergence accuracy of our proposed architecture using various characteristic functions.  Specifically, we investigate the experimental results of cognitive-only PSO, CLPSO, CHpPSO-ABS, and CHCLPSO-ABS on the CEC 2013 test suite. All tested functions are 10-dimensional, and the parameters for all algorithms are set as described in Subsection \ref{Experimental Setup}.  For a fair comparison, each algorithm uses a population size of 20 and is run 50 times independently on each test function.  The mean values of error (Mean) are presented in Table \ref{comparison results  on CEC 2013 test functions among Cognitive-Only PSO, CLPSO, CHpPSO-ABS and CHCLPSO-ABS}, with rankings shown in parentheses—red indicates first place, and blue indicates second.
	
 Comparing the results of cognitive-only PSO with  CHpPSO-ABS, we find that CHpPSO-ABS performs better on all tested functions. This improvement is due to the full utilization of constructed vectors in our architecture.  Similarly, we can still draw this conclusion by comparing CLPSO with CHCLPSO-ABS. Thus, with the help of our architecture, the performance of cognitive-only PSO and CLPSO has been improved dramatically, illustrating the effectiveness of our architecture. Moreover, we observe that CLPSO outperforms cognitive-only PSO on most functions, mainly because CLPSO's constructed vectors contain more information using the CL strategy. Furthermore, CHCLPSO-ABS consistently performs better than CHpPSO-ABS. It indicates that our proposed architecture can draw better convergence performance when the  constructed vectors contain proper information. These findings are consistent with previous results, demonstrating that our architecture can improve the convergence performance of the original algorithms. This further validates the generalization performance of our architecture.
	
		\textit{3) Analysis of the Sensitivity of Parameter:}  \label{Analysis of the Sensitivity of parameter} In this part, we investigate the sensitivity of the CH\textit{x}PSO-ABS architecture to the total upper threshold $M$. This parameter is important because it controls how long one fixed constructed vector remains in the evolutionary process. A small $M$ value may lead to the under-utilization of the constructed vector, while a large value may result in over-utilization, wasting computational resources and limiting the exploration of unknown regions. Thus, $M$ should be neither too large nor too small \cite{liang2006comprehensive}. Besides, an effective architecture should be robust across a certain range of parameter values \cite{cao2018comprehensive}. If it only works well for a specific value but poorly for others, its robustness is questionable. For this purpose, we empirically set $M$ to  $3, 4, 5, 6, 7$. We investigate  the mean values of error of CHpPSO-ABS and CHCLPSO-ABS at different values of $M$ on the CEC 2013 test suite. The results are shown in Table \ref {Mean of CHxPSO-ABS Algorithms}.
		
The results show  that there is almost no difference among the experimental results at different $M$ values for both CHpPSO-ABS and CHCLPSO-ABS, demonstrating the robustness of our architecture. Thus, we recommend $M$ within the range $[3, 7] $. In this paper, we set $ M = 6$. Additionally, we find that CHCLPSO-ABS consistently outperforms CHpPSO-ABS at different values of $M$ on most tested functions, further supporting our previous conclusions.

	\subsection{Experimental Results on  Convergence Performance of CHCLPSO-ABS Compared to Selected Algorithms} \label{Experimental Results of Performance and Generalizability}
\begin{figure*}[t]
			\centering
			\includegraphics[width=6.9in]{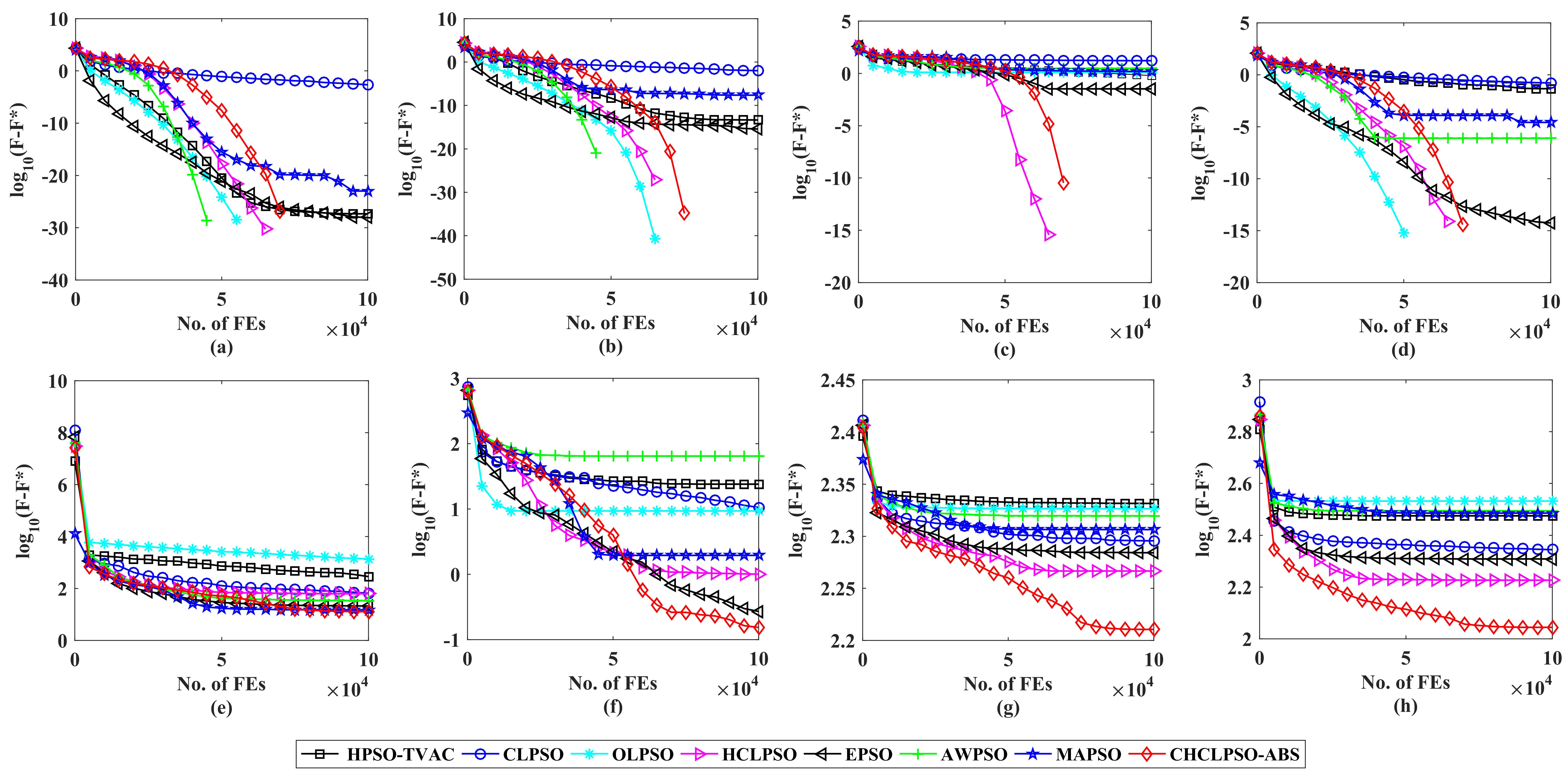}
			\caption{Performance curves of 8 algorithms. (a) F$_1$. (b) F$_5$. (c) F$_{13}$. (d) F$_{25}$. (e) F$_{33}$. (f) F$_{39}$. (g) F$_{44}$. (h) F$_{51}$.}
			\label{Performance curve of 8 algorithms}
		\end{figure*}

	In order to further validate the performance of our architecture, we compare CHCLPSO-ABS with seven  popular algorithms. For a comprehensive comparison, we use 57 functions. To ensure fairness, all algorithms have a fixed population size of 20 and the same number of function evaluations, set at $FEs_{max}=10,000\cdot D$, where \textit{D} is 10. Each algorithm is run 50 times independently on each benchmark function to obtain statistical results. These results include the mean  values of error (Mean), the standard deviations of the error values (std), the ranks of the Mean (rank), and the hypothesis test outcomes (Sign-Better / Sign-Worst), as  displayed in Table \ref{comparison results of solution accuracy on unimodal functions (F$_1$-F$_7$)}, Table \ref{comparison results of solution accuracy on simple multimodal functions (F$_{8}$-F$_{29}$)}, and Table \ref{comparison results of solution accuracy on hybrid functions (F$_{30}$-F$_{57}$) }. The  best Mean results are highlighted in red bold. In these tables,  We also calculate the average rank (Ave-rank) and report the number of best Mean (Best-Mean) and worst Mean (Worst). Moreover,  the performance curves of the eight algorithms on eight representative benchmark functions are  shown in  Fig. \ref{Performance curve of 8 algorithms}.

		\begin{table*}[!t]
	\centering
	\caption{Comparison Results  on Unimodal Functions (F$_1$-F$_7$) }
	\includegraphics[width=6.2in]{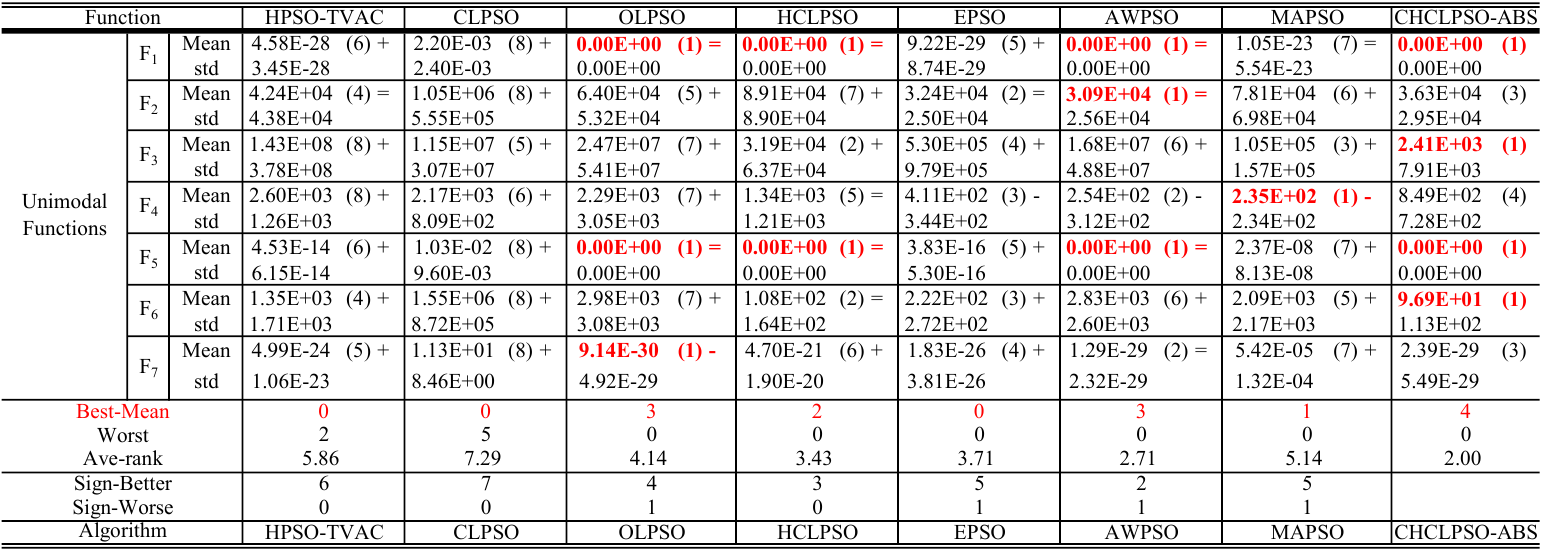}
	
	\label{comparison results of solution accuracy on unimodal functions (F$_1$-F$_7$)}
\end{table*}

\begin{table*}[!t]
	\centering
	\caption{Comparison Results on Simple Multimodal Functions (F$_{8}$-F$_{29}$) }
	\includegraphics[width=6.2in]{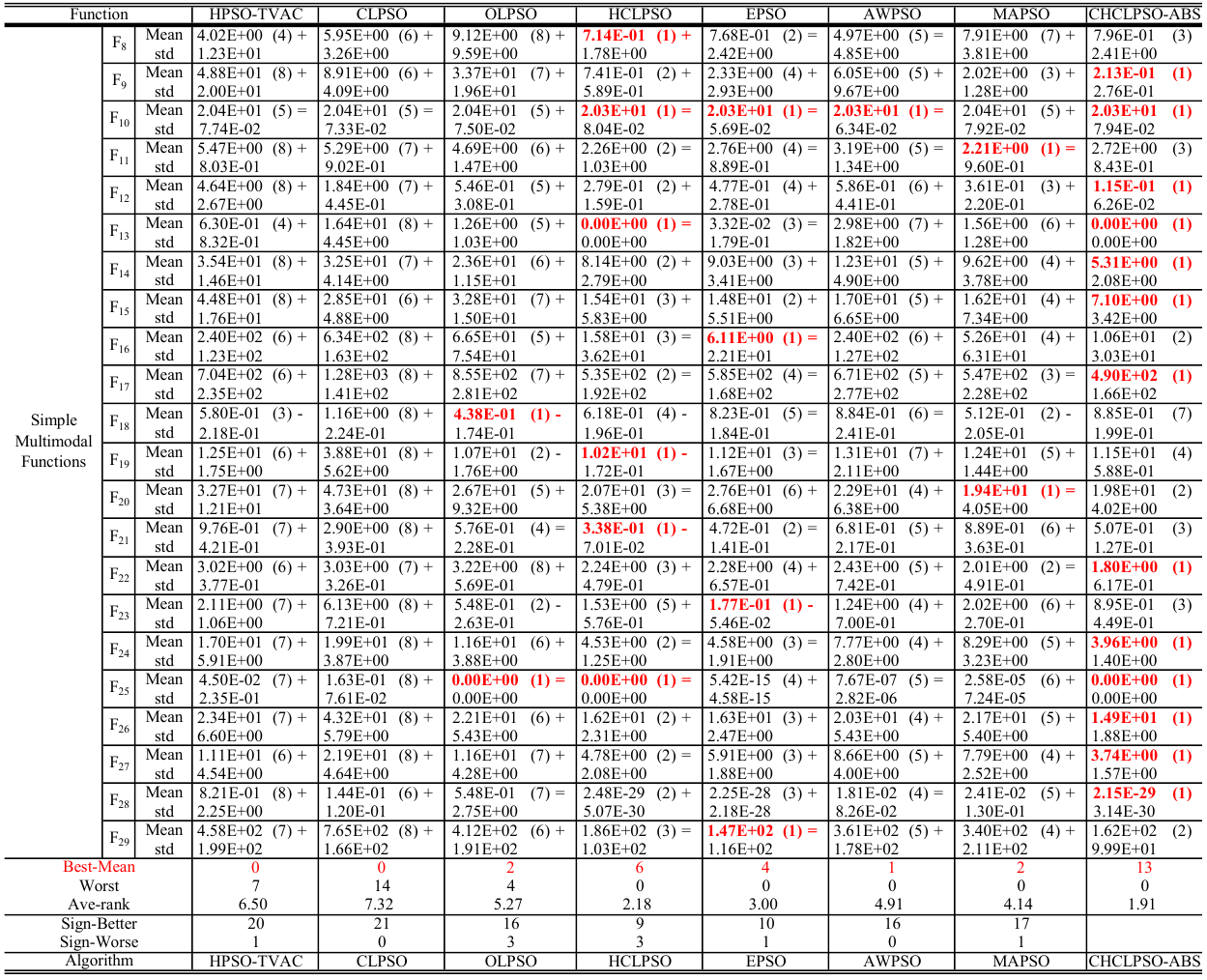}
	
	\label{comparison results of solution accuracy on simple multimodal functions (F$_{8}$-F$_{29}$)}
\end{table*}

\begin{table*}[t]
	\centering
	\caption{Comparison Results  on Complex Multimodal  Functions (F$_{30}$-F$_{57}$) }
	\includegraphics[width=6.2in]{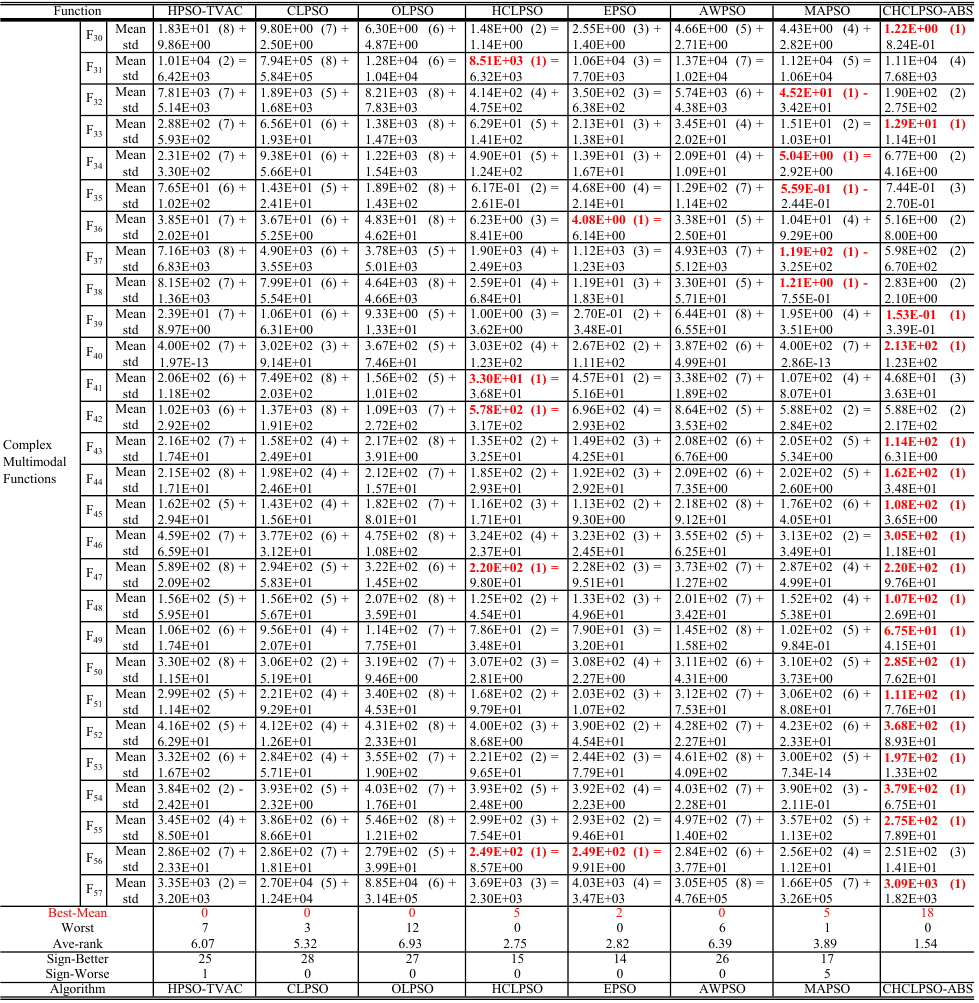}
	
	\label{comparison results of solution accuracy on hybrid  functions (F$_{30}$-F$_{57}$) }
\end{table*}

Table \ref{comparison results of solution accuracy on unimodal functions (F$_1$-F$_7$)} presents the results of the eight algorithms on Unimodal Functions (F$_1$-F$_7$).  CLPSO struggles to achieve high solution accuracy. This is expected, as CLPSO has strong global search ability but weak local convergence due to its specially designed learning strategy \cite{cao2018comprehensive}. According to the Best-Mean, OLPSO and AWPSO achieve high solution accuracy, perhaps because both algorithms effectively use \textit{\textbf{G}} to quickly aggregate particles. Although HCLPSO also uses \textit{\textbf{G}}, its convergence accuracy is slightly weaker than OLPSO and AWPSO. However, HCLPSO performs slightly better in Ave-rank, suggesting stronger generalization performance than OLPSO, mainly due to its heterogeneous update mechanisms. Our proposed algorithm performs best in both Best-Mean and Ave-rank, which is thanks to the CH\textit{x}PSO framework and the ABS strategy, which dynamically controls the proportion of particles involved in evolution in the two update mechanisms. Moreover, the hypothesis test results also confirm the significant effectiveness of our proposed algorithm, showing that our architecture  greatly enhances the convergence accuracy of CLPSO on Unimodal Functions. These results illustrate the importance of update-rule heterogeneity and accurately controlling the proportion of particles with different behaviors \cite{de2009heterogeneous}. Additionally, CHCLPSO-ABS ranks first on both F$_3$ and F$_6$, two derivative functions of the Bent Cigar Function. This indicates that our architecture has strong generalization performance when dealing with different operations on a single unimodal function. This is largely attributed to the interaction between our architecture and  the problem's fitness landscape.

The experimental results on Simple Multimodal Functions (F$_8$-F$_{29}$) are illustrated in Table \ref{comparison results of solution accuracy on simple multimodal functions (F$_{8}$-F$_{29}$)}. According to the Best-Mean,  CHCLPSO-ABS achieves the most first-rank positions, more than twice as many as HCLPSO, which ranks second.  Although OLPSO and AWPSO perform well on Unimodal Functions, they underperform on Simple Multimodal Functions. This may be due to the limitations of one single update mechanism, which seems to be consistent with the inference in \cite{de2009heterogeneous}.  Besides, HCLPSO algorithm outperforms OLPSO and AWPSO, demonstrating the advantages of heterogeneous update mechanisms. Moreover, our proposed algorithm shows significant superiority in both Ave-rank and hypothesis test results, indicating excellent convergence performance. Further considering two sets, $\{$F$_{13}$, F$_{14}$, F$_{24}$$\}$ (variants of the Rastrigin Function) and $\{$F$_{16}$, F$_{17}$, F$_{29}$$\}$ (variants of the Schwefel Function), our proposed algorithm achieves superior results compared to others. This suggests that CH\textit{x}PSO-ABS architecture exhibits strong generalization performance across various transformations (such as rotated, shifted) of simple multimodal functions.

Table \ref{comparison results of solution accuracy on hybrid functions (F$_{30}$-F$_{57}$) } presents the experimental results on Complex Multimodal Functions (F$_{30}$-F$_{57}$), which are primarily used to simulate real-world optimization problems. According to Worst, we observe that CLPSO performs relatively better on these functions than on Unimodal and Simple Multimodal Functions, which is mainly attributed to the CL strategy. Furthermore, HCLPSO, which builds on CLPSO, performs even better, ranking second in Ave-rank, which is thanks to the additionally employed update mechanism of modified \textit{global} PSO. However, due to the lack of the precise control of the proportion of particles in two update mechanisms, HCLPSO's performance falls behind our proposed CHCLPSO-ABS. Moreover, the hypothesis test results further confirm the superiority of CHCLPSO-ABS over the other seven algorithms in this group. It is also found that OLPSO and AWPSO still underperform, while MAPSO performs better on Complex Multimodal Functions than on Unimodal or Simple Multimodal ones.

Jointly considering Table \ref{comparison results of solution accuracy on unimodal functions (F$_1$-F$_7$)}, Table \ref{comparison results of solution accuracy on simple multimodal functions (F$_{8}$-F$_{29}$)} and Table \ref{comparison results of solution accuracy on hybrid functions (F$_{30}$-F$_{57}$) }, we find that as the difficulty of the functions increases, OLPSO and AWPSO's performance relatively declines, while MAPSO relatively improves. This highlights the difficulty of balancing exploration and exploitation within one single mechanism. HCLPSO performs more consistently across all three groups, demonstrating the benefits of update-rule heterogeneity. However, HCLPSO seems to be inferior compared to our proposed algorithm. This is because HCLPSO does not consider the proportion of particles involved in evolution in the two update mechanisms. With the help of the proposed ABS strategy, this gap is bridged.

	\begin{figure*} [!]
		\subfigure[]{
			
			\includegraphics[width=2.25in]{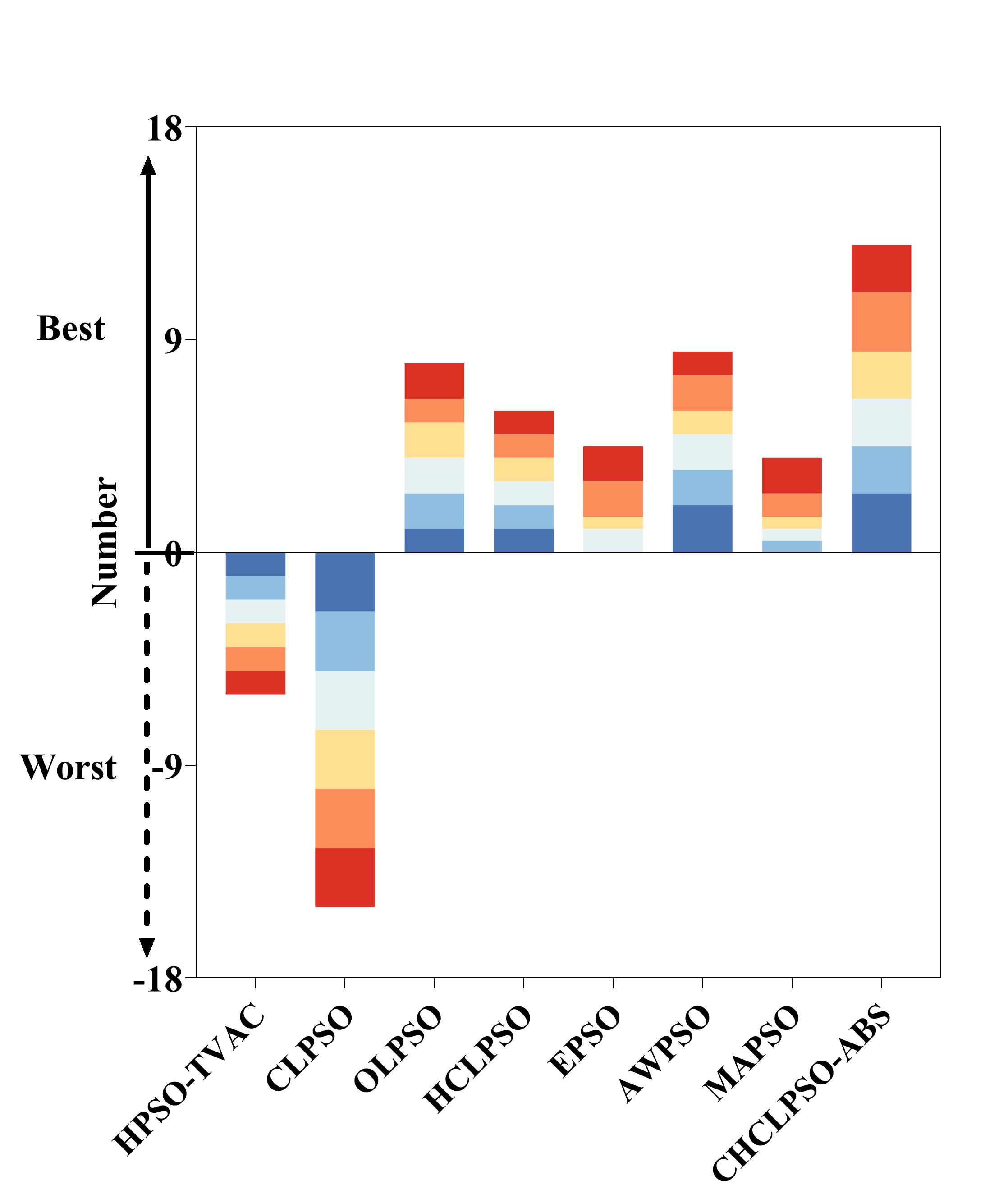}\\
		}%
		\hspace{-2mm}
		\subfigure[]{
			\includegraphics[width=2.1in]{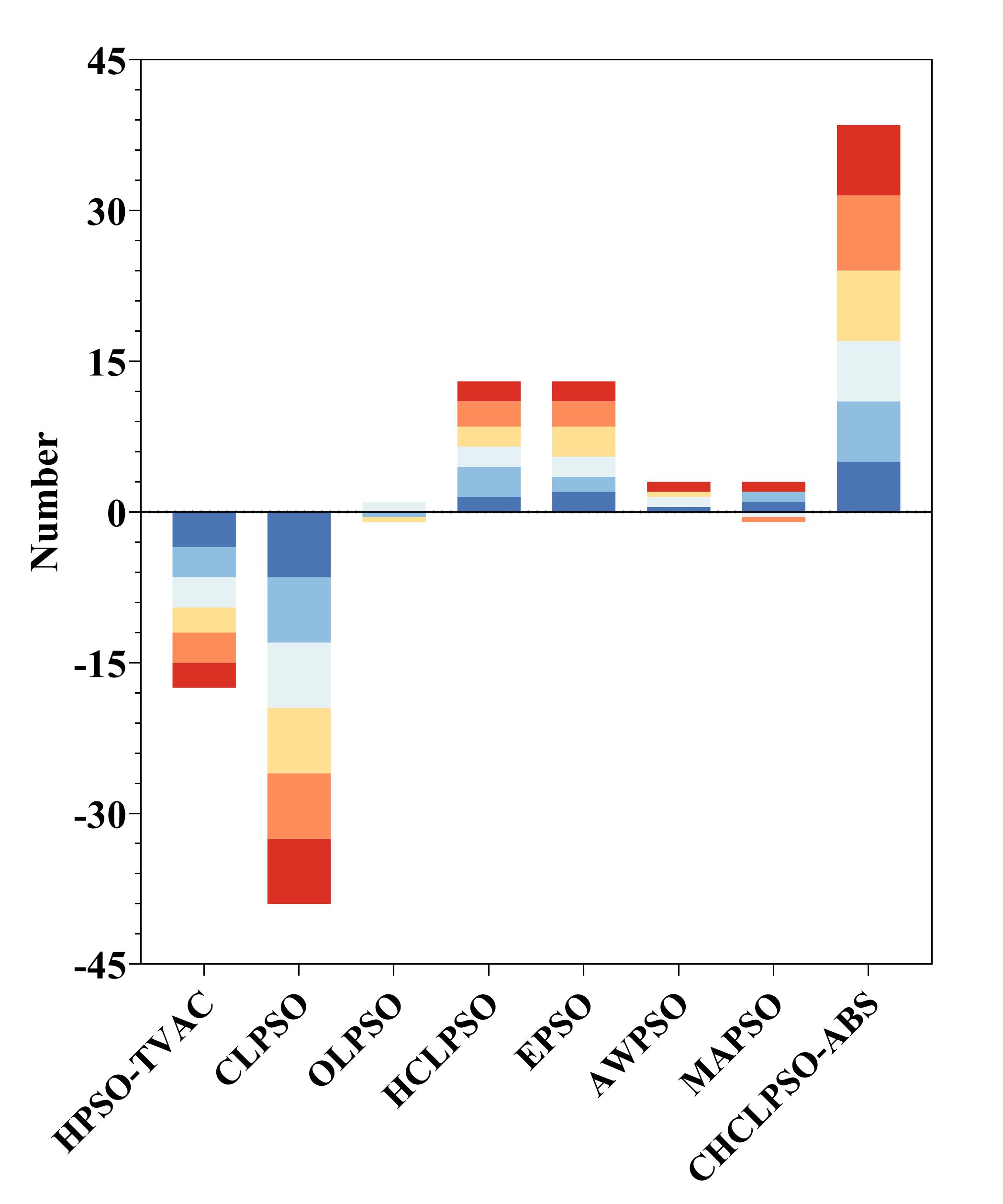}\\
		}
		\hspace{-2mm}
		\subfigure[]{
			\includegraphics[width=2.5in]{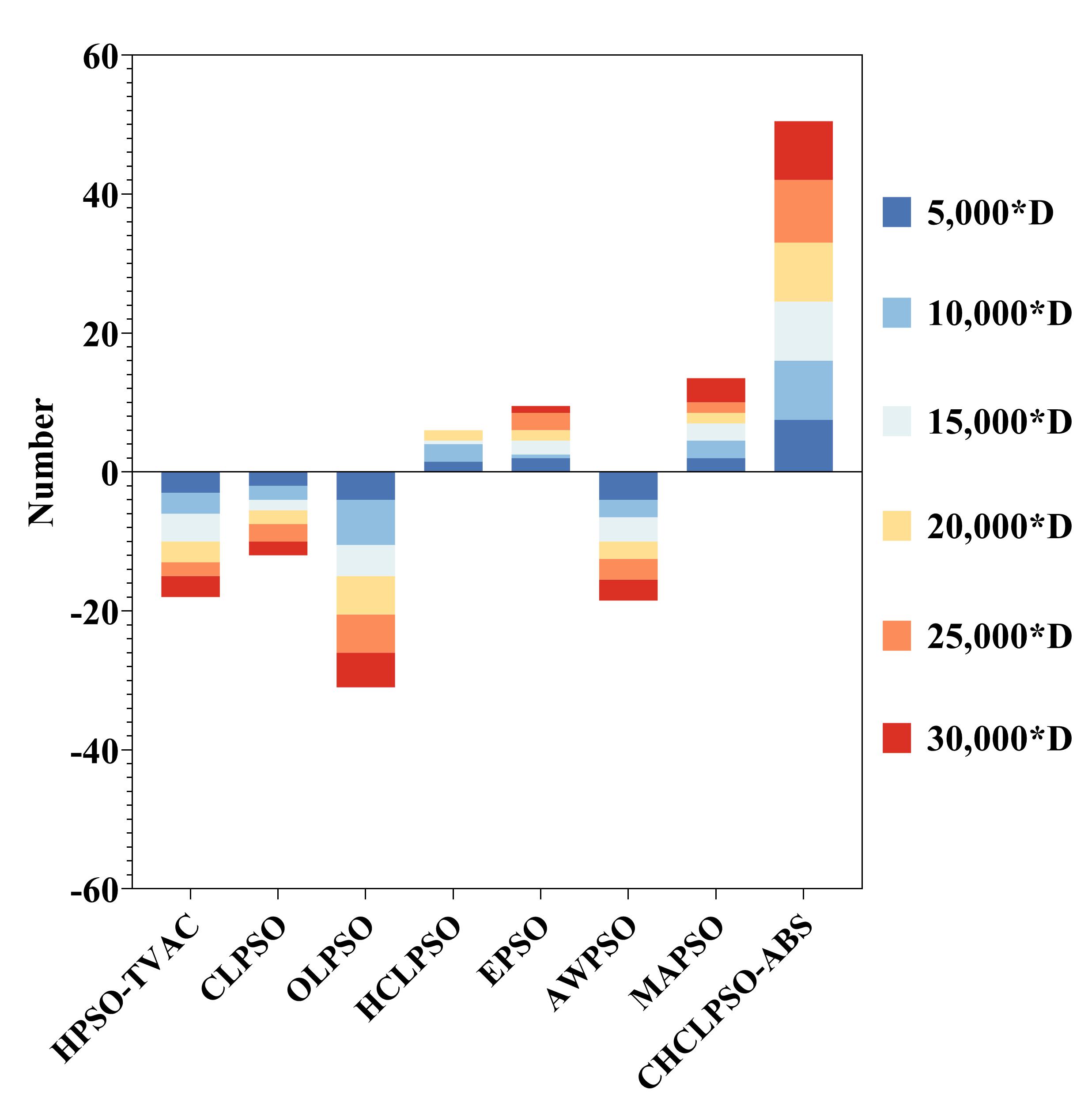}\\
		}%
		\centering
		\caption{
			Comparison results under different $FEs_{max}$. (a) Unimodal Functions.   (b) Simple Multimodal Functions. (c) Complex Multimodal Functions.}
		\label{Performance of 8 algorithms affected by $FEs_{max}$  on (a) Unimodal Functions,   (b) Simple MultimodalFunctions, (c) Hybrid Functions, (d) Composition Functions.}
	\end{figure*}

Furthermore, we illustrate the performance curves of eight algorithms on some representative functions in Fig. \ref{Performance curve of 8 algorithms}. It is observed that AWPSO shows the fastest convergence, followed by OLPSO in Fig. \ref{Performance curve of 8 algorithms}(a) and (b), while they perform worse in Fig. \ref{Performance curve of 8 algorithms}(e)-(h), confirming that they prefer local search due to one single update mechanism. Moreover, what is abrupt is that OLPSO finds the optimal solution while AWPSO does not in Fig. \ref{Performance curve of 8 algorithms}(d). It may be attributed to OLPSO's orthogonal learning strategy. Although this strategy increases diversity and  slightly slows down convergence speed, it helps find better directions\cite{zhan2009orthogonal}. This obeys the No Free Lunch Theorem \cite{wolpert1997no}. Besides, we observe that only HCLPSO and CHCLPSO-ABS find optimal solutions in Fig. \ref{Performance curve of 8 algorithms}(a)-(d). This is a significant superiority over other algorithms, benefiting from the update-rule heterogeneity. We also observe that although CHCLPSO-ABS converges slightly slower on functions where optimal solutions are found in Fig. \ref{Performance curve of 8 algorithms}(a)-(d), it shows better convergence on functions where none of the eight algorithms find the optimal solutions in Fig. \ref{Performance curve of 8 algorithms}(e)-(h). This is because our architecture maintains great diversity in the early evolutionary process and speeds the convergence in the later process to improve solution accuracy. Overall, the CHCLPSO-ABS algorithm has significant  convergence accuracy on most functions, with only a relatively slight cost in convergence speed, which is acceptable in practical applications and aligns with the No Free Lunch Theorem. In conclusion, with the help of our architecture, the optimization performance of CLPSO can be dramatically enhanced.

	\subsection{Experimental Results of Stability of CHCLPSO-ABS Compared to Selected Algorithms on  Different \textit{FEsmax} }
	
	It is a well-established fact that the performance of an algorithm is influenced by the number of $FEs_{max}$, with larger values generally improving the performance. However, the degree of performance improvement varies among different algorithms. Thus, evaluating the stability of performance across different $FEs_{max}$ values is crucial. In this section, we design six cases with $FEs_{max} = 5,000\cdot D, 10,000\cdot D, 15,000\cdot D, 20,000\cdot D, 25,000\cdot D, 30,000\cdot D$, where $D=10$. Each algorithm is tested in 30 independent experiments on 57 benchmark functions, with the number of first and last ranks counted for each case. The results are shown in Fig. \ref{Performance of 8 algorithms affected by $FEs_{max}$ on (a) Unimodal Functions, (b) Simple MultimodalFunctions, (c) Hybrid Functions, (d) Composition Functions.}. The upper bars (Best) represent the number of times that algorithms rank first, and the lower bars (Worst) show the number of times that they rank last.
	
Overall, our proposed algorithm consistently ranks the highest in each or all block heights, demonstrating its superiority. In Fig. \ref{Performance of 8 algorithms affected by $FEs_{max}$ on (a) Unimodal Functions, (b) Simple MultimodalFunctions, (c) Hybrid Functions, (d) Composition Functions.}(a), our algorithm ranks the highest across all $FEs_{max}$ cases, while CLPSO ranks last, highlighting the improvement ability of our architecture for CLPSO. Moreover, HCLPSO is outperformed by AWPSO and OLPSO, which prefer exploitation. However, the results on Simple Multimodal Functions in Fig. \ref{Performance of 8 algorithms affected by $FEs_{max}$ on (a) Unimodal Functions, (b) Simple MultimodalFunctions, (c) Hybrid Functions, (d) Composition Functions.}(b) display a rapid performance loss of OLPSO and AWPSO,  which are consistent with our results in Table \ref{comparison results of solution accuracy on simple multimodal functions (F$_{8}$-F$_{29}$)}, Table \ref{comparison results of solution accuracy on hybrid  functions (F$_{30}$-F$_{57}$) }, and Fig \ref{Performance curve of 8 algorithms}.  Moreover, in Fig. \ref{Performance of 8 algorithms affected by $FEs_{max}$ on (a) Unimodal Functions, (b) Simple MultimodalFunctions, (c) Hybrid Functions, (d) Composition Functions.}(b), OLPSO and MAPSO even occur below baseline 0 in some cases, showing their sensitivity to $FEs_{max}$.
 By contrast, CHCLPSO-ABS performs well. In Fig. \ref{Performance of 8 algorithms affected by $FEs_{max}$ on (a) Unimodal Functions, (b) Simple MultimodalFunctions, (c) Hybrid Functions, (d) Composition Functions.}(c), we also observe that OLPSO and AWPSO's performance relatively declines, while CLPSO and MAPSO show relative improvement. These results are consistent with those in Section \ref{Experimental Results of Performance and Generalizability}. However, compared to our algorithm, the other seven algorithms show large fluctuations in performance across different $FEs_{max}$ values, indicating their relatively strong reliance on $FEs_{max}$. This is because they lack explicit control over the proportion of particles in two update mechanisms and the interaction with the problem's fitness landscape. This further validates the advantages of our architecture.

	\section{Conclusion}\label{conclusion}
In this paper, we proposed a PSO architecture, CH\textit{x}PSO-ABS. This architecture employs two update channels and two corresponding subswarms to address the insufficient utilization of the two abilities of constructed vectors in modified cognitive-only PSOs. Besides, This architecture also includes two replaceable components, the information pool, and the operator, which ensure that a series of cognitive-only PSO algorithms can be embedded into our architecture. We also introduced an ABS strategy, which dynamically controls the proportion of particles in two update channels during the evolutionary process based on the number of function evaluations and the updated states of single-layer best vectors, which thereby guarantees the flexible utilization of the constructed vectors. These components work together to enhance the performance of a series of modified cognitive-only PSOs. We also embedded the vector-construction methods of cognitive-only PSO and CLPSO into our architecture, obtaining the CHpPSO-ABS and CHCLPSO-ABS, respectively.

In addition, we also experimentally demonstrated the generalization performance of our architecture, specifically the exploration and exploitation ability, the convergence accuracy performance, and the sensitivity to the total upper threshold. Besides, we also experimentally verified the convergence performance of CLPSO after embedded into our architecture and its stability to the number of function evaluations.

\clearpage
	\bibliographystyle{IEEEtran}
	\bibliography{Main_Manuscript}
	\vfill
	
\end{document}